\documentclass{article}

\usepackage[main, final]{neurips_2026}
\usepackage[utf8]{inputenc} 
\usepackage[T1]{fontenc}    
\usepackage{hyperref}       
\usepackage{url}            
\usepackage{booktabs}       
\usepackage{amsfonts}       
\usepackage{nicefrac}       
\usepackage{microtype}      
\usepackage{xcolor}         
\usepackage{subcaption}
\usepackage{makecell}
\usepackage{wrapfig}
\usepackage{graphicx}
\usepackage{amsmath}
\usepackage{amssymb}
\usepackage{listings}
\usepackage{enumitem}
\setlist[itemize]{align=parleft,left=0pt..1em}

\title{X2HDR: HDR Image Generation in a Perceptually Uniform Space}

\author{
	Ronghuan Wu$^{1}$, \
	Wanchao Su$^{2}$, \
	Kede Ma$^{1}$, \
	Jing Liao$^{1}$,
	and
	Rafał K. Mantiuk$^{3}$
	\\
	$^1$City University of Hong Kong \\
	$^2$Monash University \\
	$^3$University of Cambridge \\
	{\tt\small rh.wu@my.cityu.edu.hk}, 
	{\tt\small wanchao.su@monash.edu}, 
	{\tt\small kede.ma@cityu.edu.hk}, \\
	{\tt\small jingliao@cityu.edu.hk}, 
	{\tt\small rafal.mantiuk@cl.cam.ac.uk}
}

\newcommand{\sysName}{X2HDR}
\newcommand{\cdms}{\,cd/m$^2$}
\newcommand{\ie}{\textit{i}.\textit{e}.}
\newcommand{\eg}{\textit{e}.\textit{g}.}
\def\etal{\emph{et al.}}

\definecolor{codegreen}{rgb}{0,0.6,0}
\definecolor{codegray}{rgb}{0.5,0.5,0.5}
\definecolor{codepurple}{rgb}{0.58,0,0.82}
\definecolor{backcolour}{rgb}{0.961,0.984,0.984}
\lstdefinestyle{mystyle}{
  backgroundcolor=\color{backcolour},
  commentstyle=\color{codegreen},
  keywordstyle=\color{magenta},
  stringstyle=\color{codepurple},
  basicstyle=\ttfamily\footnotesize,
  breakatwhitespace=false,
  breaklines=true,
  captionpos=b,
  keepspaces=true,
  numbers=none,
  showspaces=false,
  showstringspaces=false,
  showtabs=false,
  tabsize=2,
  frame=ltb,
  framerule=0pt,
}
\begin{document}

\maketitle

\begin{figure}[h]
  \vspace{-2em}
  \centering
  \includegraphics[width=\textwidth]{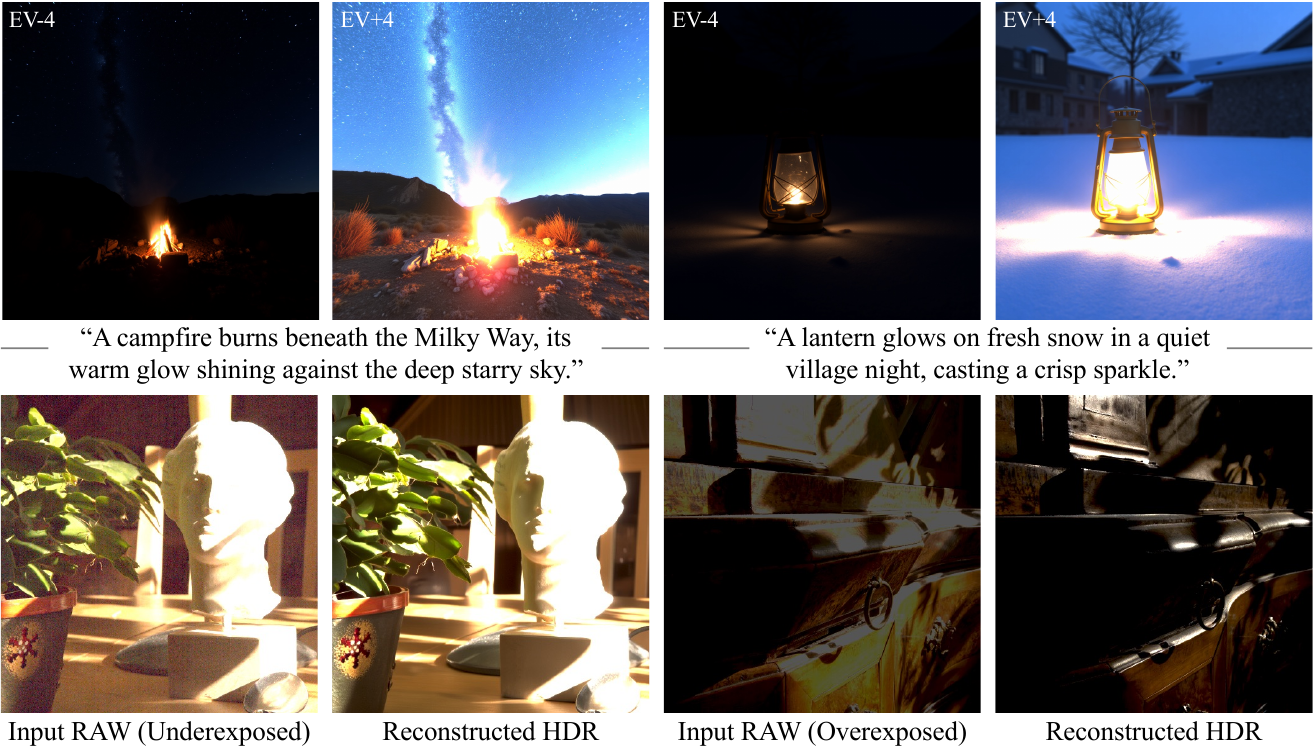}
  \caption{Qualitative results demonstrating the two modes supported by the proposed \sysName. Top: text-to-HDR generation, visualized with exposure-adjusted views at EV $-4$ and EV $+4$, highlighting synthesized content across highlights and shadows. Bottom: single-image RAW-to-HDR reconstruction. From underexposed and overexposed RAW inputs, \sysName\ reconstructs HDR structures by inpainting saturated regions and denoising in low-light areas.
  }
  \label{fig:teaser}
\end{figure}

\begin{abstract}
  High-dynamic-range (HDR) formats and displays are becoming increasingly prevalent, yet state-of-the-art image generators (\eg, Stable Diffusion and FLUX) typically remain limited to low-dynamic-range (LDR) output due to the lack of large-scale HDR training data. In this work, we show that existing pretrained diffusion models can be easily adapted to HDR generation without retraining from scratch. A key challenge is that HDR images are natively represented in linear RGB, whose intensity and color statistics differ substantially from those of sRGB-encoded LDR images. This gap, however, can be effectively bridged by converting HDR inputs into perceptually uniform encodings (\eg, using PU21 or PQ). Empirically, we find that LDR-pretrained variational autoencoders (VAEs) reconstruct PU21-encoded HDR inputs with fidelity comparable to LDR data, whereas linear RGB inputs cause severe degradations. Motivated by this finding, we describe an efficient adaptation strategy that freezes the VAE and finetunes only the denoiser via low-rank adaptation in a perceptually uniform space. This results in a unified computational method that supports both text-to-HDR synthesis and single-image RAW-to-HDR reconstruction. Experiments demonstrate that our perceptually encoded adaptation consistently improves perceptual fidelity, text-image alignment, and effective dynamic range, relative to previous techniques.
  Complete HDR results and code are available at {\color{blue}\url{https://X2HDR.github.io/}}.
\end{abstract}


\section{Introduction}
High-dynamic-range (HDR) imagery provides a more faithful representation of natural scene luminance and color than conventional low-dynamic-range (LDR) formats, offering a more realistic and appealing visual appearance. 
Yet, creating HDR content in practice is still inconvenient. The standard solution---multi-exposure bracketing---often suffers from motion-induced misalignment and ghosting, and it increases both capture and processing cost.
Specialized HDR sensors can mitigate these issues, but remain expensive and are not widely available.

In parallel, recent advances in text-to-image (T2I) diffusion models have made visual content creation accessible from simple inputs such as text prompts and coarse visual guidance (\eg, scribbles, drafts, or low-resolution images).
Extending this convenience to the HDR domain is highly desirable: users should be able to create HDR content directly from text or from a low-quality camera input, without relying on bracketing software or specialized hardware.

Several recent methods attempt to adapt pretrained T2I models to support HDR by emulating the classic HDR pipeline: generate multi-exposure ``brackets'' and then merge them into an HDR output~\cite{debevec2023recovering}. While effective in principle, this strategy introduces substantial architectural and algorithmic complexity. For example, LEDiff~\cite{wang2025lediff} trains separate denoisers for highlight hallucination and shadow recovery, and additionally finetunes the variational autoencoder (VAE) decoder~\cite{kingma2013auto} to output linear HDR values. Bracket Diffusion~\cite{bemana2025bracket} jointly denoises multi-exposure latents, incurring inference time and memory costs that scale with the number of brackets. These design choices complicate deployment and hinder adoption of newer, more memory-intensive backbone architectures.

We argue that much of this complexity stems from a simpler root cause: a representational mismatch between LDR and HDR imagery. Most contemporary diffusion models are trained on billions of display-encoded, nonlinearly compressed LDR images, whereas HDR and RAW data are natively expressed in a linear-light space prior to the image signal processor (ISP), resulting in dramatically different pixel-intensity statistics. In particular, human visual sensitivity to luminance changes is much higher in shadows than in highlights~\cite{mantiuk2004perception}, making the linear HDR distribution poorly aligned with the perceptually shaped LDR data.

Building on this insight, we present \sysName, a unified computational method for HDR image synthesis and reconstruction with minimal changes to existing T2I models. The key to \sysName\ is to operate in a perceptually uniform space (\eg, induced by PU21~\cite{Mantiuk2021} or PQ~\cite{miller2013perceptual}).
Specifically, we first map HDR data into this perceptual space, which compresses extreme highlights and reallocates precision toward shadows, in a way that reshapes HDR luminance statistics to better match LDR pretraining data. Empirically, this simple form of preprocessing empowers LDR-pretrained VAEs to recover PU21-encoded HDR inputs with fidelity comparable to LDR reconstructions, whereas linear HDR leads to distorted latents and severe artifacts. Leveraging this, \sysName\ freezes the VAE and acquires HDR capability by finetuning only the denoiser through parameter-efficient low-rank adaptation (LoRA)~\cite{hu2021lora}.

We instantiate \sysName\ on two representative tasks:
1) text-to-HDR generation, synthesizing HDR images directly from text, and
2) single-image RAW-to-HDR reconstruction, in which the strong generative prior of the pretrained T2I model facilitates plausible hallucination of missing content in overexposed regions while suppressing noise in underexposed areas.
Qualitative and quantitative evaluations, together with a formal perceptual study (see Supplemental Section~\ref{sec_supp:perceptual_study}) on a calibrated HDR display, show that the proposed \sysName\ consistently improves perceptual fidelity, text-image alignment, and effective dynamic range over prior approaches (see Figure~\ref{fig:teaser}).

In summary, our main contributions are
\begin{itemize}
    \item A simple yet effective practice for HDR adaptation---encoding HDR/RAW inputs in a perceptually uniform space---which avoids complex bracket-and-merge pipelines and enables pretrained T2I models to support HDR with minimal modification;
    \item A unified computational method for text-to-HDR generation and RAW-to-HDR reconstruction with substantial perceptual improvements over previous methods.
\end{itemize}

\section{Perceptually Uniform HDR Representation}
\label{sec:method}
In this section, we present \sysName's PU21-based HDR representation~\cite{Mantiuk2021}, supporting faithful reconstruction with an LDR-pretrained VAE.

\subsection{HDR Image Representation}
\label{sec:method_hdr_representation}
HDR (and RAW) images are typically stored in a linear color space, where pixel values are proportional to the underlying scene/sensor light signal up to a scale factor.
However, linear representation is not perceptually uniform: an equal luminance increment is far more noticeable at low levels than at high levels. Perceptually uniform encodings address this mismatch by redistributing encoded values according to human visual sensitivity, including SMPTE PQ~\cite{miller2013perceptual} and PU21~\cite{Mantiuk2021}.
For example, under PQ, an increase of $1$\cdms{} near darkness ($0.005$\cdms{}) is over $150{\times}$ more salient than the same increase at $100$\cdms{}.
Beyond perceptual motivation, prior work~\cite{ke2023training} also shows that networks train more effectively on PQ/PU21-encoded HDR/RAW data, rather than represented in linear space. These observations suggest adopting a perceptually uniform representation for HDR synthesis.

In \sysName, we encode linear HDR values using PU21, approximated by a log-quadratic function~\cite{ke2023training}:
\begin{equation}
\label{eq:pu21}
    V= f_{\mathrm{\,PU21}}(L) \;=\; a\,\bigl(\log_2(L) - L_{\min}\bigr)^2 \;+\; b\,\bigl(\log_2(L) - L_{\min}\bigr),
\end{equation}
where $L$ denotes an absolute linear RGB value within $[0.005,10,000]$. The fitted parameters are $a=0.001908$ and $b=0.0078$, and $L_{\min}$ is set to $\log_2(0.005)$.
PU21 compresses extreme highlights while allocating more representational resolution to low-luminance values (see Supplemental Section~\ref{sec_supp:encoding_functions}).
The inverse transform is
\begin{equation}
\label{eq:inverse_pu21}
   L= f_{\mathrm{PU21}}^{-1}(V) \;=\; 2^{\frac{2aL_{\min}-b+\sqrt{\,b^2+4aV\,}}{2a}}.
\end{equation}

For all HDR images, we globally rescale linear RGB values so that the maximum luminance corresponds to $L_{\mathrm{peak}}=4,000$\cdms{} (matching the peak luminance of commercially available HDR displays). We then apply $f_{\,\mathrm{PU21}}(\cdot)$ channel-wise to map the rescaled values to $[0,1]$. For generated outputs, we apply $f_{\mathrm{PU21}}^{-1}(\cdot)$ to recover linear RGB values.

\subsection{Pretrained VAEs for HDR Reconstruction}
\label{sec:method_vae_recon}

Modern diffusion models typically employ a VAE to map LDR images into a compact latent space, while reconstructing them with high fidelity. This raises a practical question central to \sysName: must the VAE be finetuned for HDR reconstruction, or can an LDR-pretrained one already reconstruct HDR content accurately, provided an appropriate encoding is used?

\paragraph{Setups}
We curate a Blu-ray movie (namely, ``BBC Blue Planet II Episode 2 - The Deep'') that provides both LDR and HDR versions of the same content.
Frames are time-synchronized by selecting the temporal offset that maximizes cross-correlation between corresponding pixel values.
After temporal alignment and resolution matching, we randomly crop and resize paired frames to $768{\times}768$, producing a set of $512$ (LDR, HDR) pairs with identical scene content.
LDR frames are rescaled to $[0,1]$, while HDR frames are processed in two ways: 1) converted into a perceptually uniform space using Equation~(\ref{eq:pu21}), or 2) kept in linear space but normalized to $[0,1]$.
We then encode and decode all images using the pretrained \texttt{FLUX.1-dev} VAE. Reconstruction fidelity is evaluated by ColorVideoVDP~\cite{Mantiuk_2024_cvvdp}, which reports perceptual just-objectionable-difference (JOD) scores (one JOD corresponds to roughly a $75\%$ observer preference), together with standard LDR metrics~\cite{zhang2018unreasonable, wang2004image} applied via exposure-optimized inverse display models~\cite{cao2024perceptual}.

\begin{table}[t]
    \centering
    \footnotesize
    \setlength{\tabcolsep}{4pt}
    \caption{
    VAE reconstruction fidelity on $512$ aligned (LDR, HDR) frame pairs. The JOD score is given by ColorVideoVDP~\cite{Mantiuk_2024_cvvdp}. $Q^\star$ represents a family of LDR quality measures, equipped with exposure-optimized inverse display models~\cite{cao2024perceptual}.
    }
    \begin{tabular}{lccccc}
        \toprule
        Setting
        & JOD $\uparrow$ 
        & $Q^\star_{\mathrm{PSNR}} \uparrow$
        & $Q^\star_{\mathrm{SSIM}} \uparrow$
        & $Q^\star_{\mathrm{LPIPS}} \downarrow$
        & $Q^\star_{\mathrm{DISTS}} \downarrow$ \\
        \midrule
        LDR & $\mathbf{9.86}$ & $\mathbf{38.8}$ & $0.961$ & $0.026$ & $\mathbf{0.008}$ \\
        HDR (Linear) & $8.54$ & $27.9$ & $0.867$ & $0.129$ & $0.067$ \\
        HDR (PU21) & $9.44$ & $35.6$ & $\mathbf{0.965}$ & $\mathbf{0.020}$ & $0.010$ \\
        \bottomrule
    \end{tabular}
    \vspace{-0.5em}
    \label{tab:vae_recon_metrics}
\end{table}

\begin{figure}[t]
    \centering
    \begin{minipage}[c]{0.58\textwidth}
        \centering
        \includegraphics[width=\textwidth]{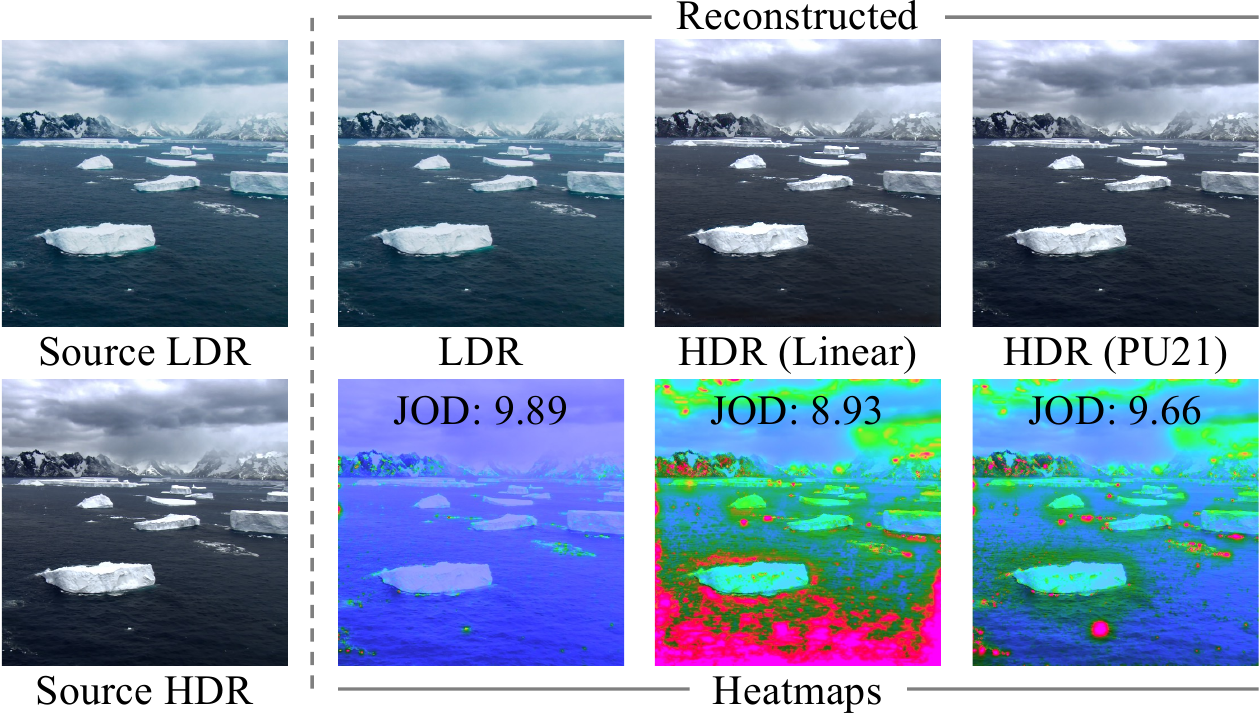}
    \end{minipage}%
    \hfill
    \begin{minipage}[c]{0.4\textwidth}
        \captionof{figure}{
        VAE reconstruction of an aligned (LDR, HDR) pair under different input encodings. Left: source LDR/HDR images fed to the VAE. Right: reconstructions (top row) for LDR, linear HDR, and PU21-encoded HDR representation, together with ColorVideoVDP perceptual error heatmaps (bottom row), where stronger colors indicate more noticeable differences. Insets report the corresponding JOD scores.
        }
        \label{fig:vae_recon}
    \end{minipage}
    \vspace{-1em}
\end{figure}

\paragraph{Results}
As summarized in Table~\ref{tab:vae_recon_metrics}, PU21-encoded HDR inputs are reconstructed nearly as well as LDR inputs: the JOD score drops only slightly ($9.86 \rightarrow 9.44$), and the other metrics show similarly small gaps. In contrast, linear HDR inputs exhibit substantial degradation across all metrics. These quantitative trends are consistent with Figure~\ref{fig:vae_recon}: LDR and PU21 reconstructions preserve scene structure and texture with only minor (and imperceptible) artifacts, whereas linear HDR reconstructions present widespread distortions.

\paragraph{Analysis}
PU21 improves VAE reconstruction by mapping HDR values to a perceptually meaningful representation that more closely matches the statistics of the LDR pretraining data.
To illustrate this effect, we construct the inset histograms from $512$ aligned LDR-HDR image pairs.
We average the three RGB channels at each pixel, pool the resulting values across all images, and plot their probability densities on a logarithmic $y$-axis.
Under linear HDR encoding, values are concentrated near zero, leaving much of the normalized input range underutilized.
This highly skewed distribution differs substantially from that of the LDR training data.
\setlength{\intextsep}{5pt}
\setlength{\columnsep}{8pt}
\begin{wrapfigure}{r}{0.4\linewidth}
    \vspace{-1pt}
    \centering
    \includegraphics[width=\linewidth]{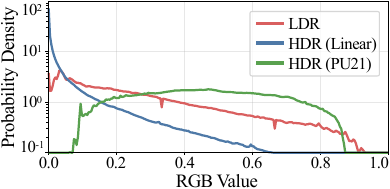}
    \vspace{-15pt}
\end{wrapfigure}
The mismatch also extends to the VAE latent space: linear HDR inputs produce a latent distribution that deviates from the LDR latent prior, with a Kullback-Leibler (KL) divergence of $0.231$ across the $512$ pairs.
By compressing extreme highlights and allocating greater representational resolution to shadows and midtones, PU21 distributes HDR values more broadly across the normalized range and reduces their concentration near zero.
This transformation brings the inputs into closer alignment with the statistics expected by the LDR-pretrained VAE, reducing the latent-space KL divergence to $0.033$.
Additional analysis and a t-SNE visualization of the VAE latent space are provided in Supplemental Sections~\ref{sec_supp:pu21_explanation} and~\ref{sec_supp:more_vae_experiments}.

\section{Text-to-HDR Image Generation}
\label{sec:text2hdr}
\begin{figure}[t]
    \centering
    \includegraphics[width=0.75\linewidth]{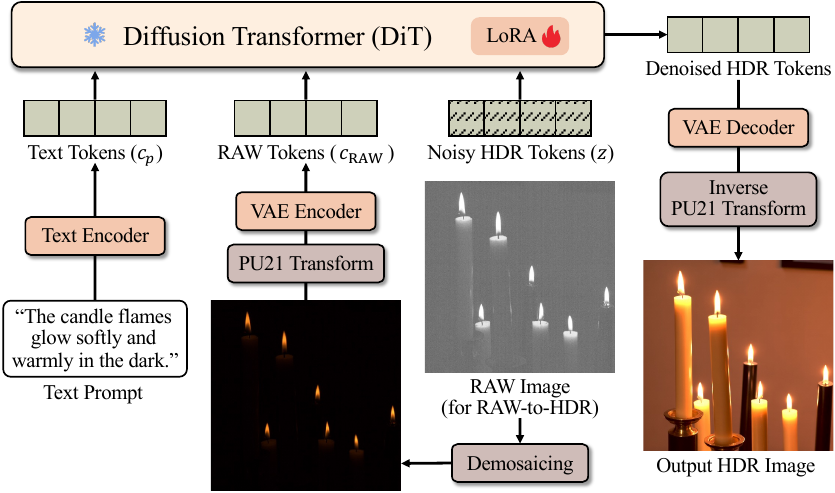}
    \caption{
    System diagram of \sysName. For \textit{Text-to-HDR}, the prompt is encoded into text tokens $c_p$;  a noisy HDR latent $z$ is sampled and denoised by the LoRA-adapted DiT; the final result is decoded by the frozen VAE and mapped back to linear HDR values via the inverse PU21 transform (Equation~(\ref{eq:inverse_pu21})). For \textit{RAW-to-HDR}, a RAW capture is demosaicked, PU21-encoded (Equation~(\ref{eq:pu21})), and converted into image tokens $c_{\mathrm{RAW}}$; conditioning on $c_{\mathrm{RAW}}$, the same denoising and decoding pipeline reconstructs the HDR output.
    Both branches are shown in a single diagram for simplicity; in practice, training is performed separately with a task-specific LoRA for each branch.
    }
\label{fig:pipeline}
\end{figure}

As demonstrated in Section~\ref{sec:method}, the pretrained VAE can faithfully encode and decode HDR images after PU21 mapping, achieving reconstruction fidelity close to that obtained on standard LDR images. This observation simplifies text-to-HDR image generation: we keep the text encoder and VAE fixed, and finetune only the denoiser (through LoRA) while operating entirely in a perceptually uniform space. Our system diagram is shown in Figure~\ref{fig:pipeline}.

\paragraph{Training Data Preprocessing}
For each training HDR image $I_{\mathrm{HDR}}$, we first apply a global rescaling so that its maximum luminance corresponds to $L_{\mathrm{peak}} = 4,000$\cdms{}, and then map the rescaled values into the PU21 space. This conversion is the only departure from standard LDR finetuning; all subsequent steps follow the standard latent generative training procedure.

\paragraph{Latent Formulation and Objective}
We map the PU21-encoded HDR image into VAE latents $x_0$ and tokenize the paired text prompt into $c_p$. We then adopt the \texttt{FLUX.1-dev} T2I backbone and finetune it with flow matching. Specifically, 
for a timestep $t \in [0,1]$, we sample $\epsilon \sim \mathcal{N}(0, I)$ and construct an interpolated noisy latent
$z = (1 - t)x_0 + t\epsilon$.
We then finetune the model by optimizing the flow-matching objective~\cite{albergo2022building, lipman2022flow, esser2024scaling}:
\begin{equation}
\ell_\mathrm{flow}\left(\Theta;c_p,x_0\right) = \mathbb{E}_{t, \epsilon} \, \left\| v_{\Theta}\left(z, t, c_p\right) - (\epsilon - x_0) \right\|_2^2,
\label{eq:loss_flow_matching}
\end{equation}
where $v_{\Theta}(\cdot)$ denotes the learned velocity field with trainable parameters $\Theta$, implemented via LoRA injection into the backbone.

\paragraph{LoRA Finetuning}
We employ LoRA~\cite{hu2021lora}, a parameter-efficient finetuning method to adapt the pretrained denoiser to the PU21-encoded HDR representation.
LoRA represents the weight update of a linear layer as a low-rank decomposition: for a pretrained weight matrix $W\in\mathbb{R}^{d_\text{out}\times d_\text{in}}$,
instead of directly optimizing $W$,
we freeze it and learn an additive update $\Delta W = BA$ with rank $r$, where $A\in\mathbb{R}^{r\times d_\text{in}}$ and $B\in\mathbb{R}^{d_\text{out}\times r}$. 
The adapted layer becomes
\begin{equation}
W' = W + \frac{\alpha}{r}BA,
\end{equation}
where $\alpha$ is a scaling factor.
We insert LoRA modules into linear layers of attention blocks (\eg, query, key, value, and output projections) and optimize only the LoRA parameters while keeping the text encoder and VAE frozen. 
This yields an HDR-capable model with minimal trainable parameters and unchanged inference architecture.

\paragraph{Inference}
At test time, given a text prompt, we sample an initial noise latent and integrate the learned flow (with merged low-rank updates) to generate a PU21-encoded HDR latent. The final HDR image in linear space is produced by the frozen VAE decoding and the inverse PU21 transform.

\section{RAW-to-HDR Image Reconstruction}
\label{sec:method_raw2hdr}

We next show that the same paradigm extends naturally to HDR reconstruction from a single RAW capture.
Unlike sRGB-encoded LDR images, RAW offers a physically meaningful (albeit noisy) linear measurement that is approximately proportional to the incident light at the sensor before the ISP's nonlinear stages of processing.

We cast RAW-to-HDR as a direct mapping from a single RAW image to an HDR output, simultaneously denoising the sensor measurement and plausibly inpainting saturated or clipped regions. Although our method could be applied to enhance sRGB-encoded LDR images (see Supplemental Section~\ref{sec_supp:ldr2hdr} for experiments in this setting), we focus on RAW inputs because 1) they typically retain more information in both underexposed and overexposed areas, and 2) they preserve approximately linear sensor values, avoiding the ill-posed problem of inverting in-camera processing. Accordingly, our goal is integration with the camera ISP rather than post-processing legacy LDR content.

Given a RAW input, we first demosaic it following Ramanath~\etal~\cite{ramanath2002demosaicking}. We then rescale the peak luminance of both the RAW input and the ground-truth (GT) HDR target to $L_{\mathrm{peak}} = 4,000$\cdms{}, and convert both to the PU21 space.
We next encode the text prompt, RAW input, and target HDR to $c_p$, $c_{\mathrm{RAW}}$, and $x_0$, respectively.
Training is guided by the same flow-matching objective $\ell_\mathrm{flow}$ with LoRA finetuning as in the text-to-HDR setting, except that the learned velocity field additionally conditions on the RAW input:
\begin{equation}
\ell_\mathrm{flow}\left(\Theta;c_p,c_\mathrm{RAW},x_0\right) = \mathbb{E}_{t, \epsilon} \, \left\| v_{\Theta}\left(z, t, c_p, c_{\mathrm{RAW}}\right) - (\epsilon - x_0) \right\|_2^2.
\end{equation}
Because our \texttt{FLUX.1-dev} backbone is DiT-based~\cite{peebles2023scalable}, supporting variable-length token sequences, we inject RAW conditioning by simply concatenating the RAW image tokens with the text and HDR latent tokens~\cite{zhang2025easycontrol, tan2025ominicontrol}. Empirically, this simple design converges fast and reliably: the model learns to recover coarse scene structure and texture early in training, while dark-region denoising and bright-region inpainting improve steadily with continued optimization.

\section{Experiments}\label{sec:exp}
In this section, we benchmark \sysName\ against previous HDR adaptation techniques that follow a bracket-and-merge paradigm (\ie, LEDiff~\cite{wang2025lediff} and Bracket Diffusion~\cite{bemana2025bracket}) and a dedicated RAW-to-HDR reconstruction method (\ie, RawHDR~\cite{zou2023rawhdr}) using metrics that capture 1) perceptual image quality and text-image alignment, 2) effective dynamic range, and 3) HDR reconstruction fidelity.

\subsection{Evaluation Metrics}
\label{sec:exp_evaluation_metrics}
\begin{itemize}
    \item \textit{Perceptual image quality and text-image alignment}. 
    We employ Q-Eval-100K~\cite{zhang2025q}, which provides two finetuned vision and language models (VLMs) (based on \texttt{Qwen2-VL-7B-Instruct}) that produce scalar scores in $[0,1]$ for perceptual image quality and text-image alignment assessment for LDR images. To better match the VLMs' expected input statistics, HDR images are first PU21-encoded prior to evaluation. A dedicated sanity check is reported in Supplemental Section~\ref{sec_supp:q_eval}.
    
    \item \textit{Effective dynamic range}. For each HDR image generated in text-to-HDR, we compute an effective dynamic-range estimate in exposure ``stops.'' We first apply Gaussian smoothing to suppress prediction noise,  $\tilde{L} = G_{\sigma} * L$, using $\sigma = 3$ pixels. We then remove extreme outliers by retaining only the $[0.5,99.5]$ percentile range, and define the dynamic range (in stops) as
    \begin{equation}\label{eq:edr}
        \mathrm{DR}_{\text{stops}} = \log_2(\tilde{L}_{99.5} / \tilde{L}_{0.5}),
    \end{equation}
    where $\tilde{L}_{0.5}$ and $\tilde{L}_{99.5}$ are the $0.5^{\mathrm{th}}$ and $99.5^{\mathrm{th}}$ percentiles of the filtered luminance, respectively.

    \item \textit{HDR reconstruction fidelity}.  For reconstruction evaluation, we use two complementary measures. We report JOD scores using ColorVideoVDP~\cite{Mantiuk_2024_cvvdp}. Meanwhile, we compute exposure-optimized variants of standard LDR metrics---PSNR, SSIM~\cite{wang2004image}, LPIPS~\cite{zhang2018unreasonable}, and DISTS~\cite{ding2020image}---via an inverse display model~\cite{cao2024perceptual} that decomposes both the reconstruction and reference into aligned exposure stacks and evaluates each LDR metric at the best-matching exposures. 
\end{itemize}

\begin{table}[t]
  \centering
  \footnotesize
  \setlength{\tabcolsep}{4pt}
  \caption{
  Quantitative results for text-to-HDR.
  We report image quality and text-image alignment scores by Q-Eval-100K~\cite{zhang2025q} (higher is better), effective dynamic range by Equation~(\ref{eq:edr}) in stops (higher is better), and inference cost measured by runtime and peak memory (lower is better).
  }
  \label{tab:text_to_hdr}
  \begin{tabular}{lccc|cc}
    \toprule
    Method
    & \begin{tabular}[c]{@{}c@{}}Image\\Quality\end{tabular} $\uparrow$
    & \begin{tabular}[c]{@{}c@{}}Text-Image\\Alignment\end{tabular} $\uparrow$
    & \begin{tabular}[c]{@{}c@{}}Dynamic\\Range\end{tabular} $\uparrow$
    & \begin{tabular}[c]{@{}c@{}}Time\\(sec)\end{tabular} $\downarrow$
    & \begin{tabular}[c]{@{}c@{}}Memory\\(GB)\end{tabular} $\downarrow$ \\
    \midrule
    LEDiff
    & $0.488$ & $0.623$ & $4.3$  & $7.4$   & $8.6$ \\
    Bracket Diffusion
    & $0.524$ & $0.671$ & $11.1$ & $557.5$ & $38.0$ \\
    Ours (\texttt{SD})
    & $0.568$ & $0.672$ & $\mathbf{14.4}$ & $\mathbf{2.2}$ & $\mathbf{2.9}$ \\
    Ours (\texttt{FLUX})
    & $\mathbf{0.668}$ & $\mathbf{0.831}$ & $14.2$ & $11.2$ & $33.0$ \\
    \bottomrule
  \end{tabular}
  \vspace{-1em}
\end{table}

\begin{figure}[t]
    \centering
    \includegraphics[width=\textwidth]{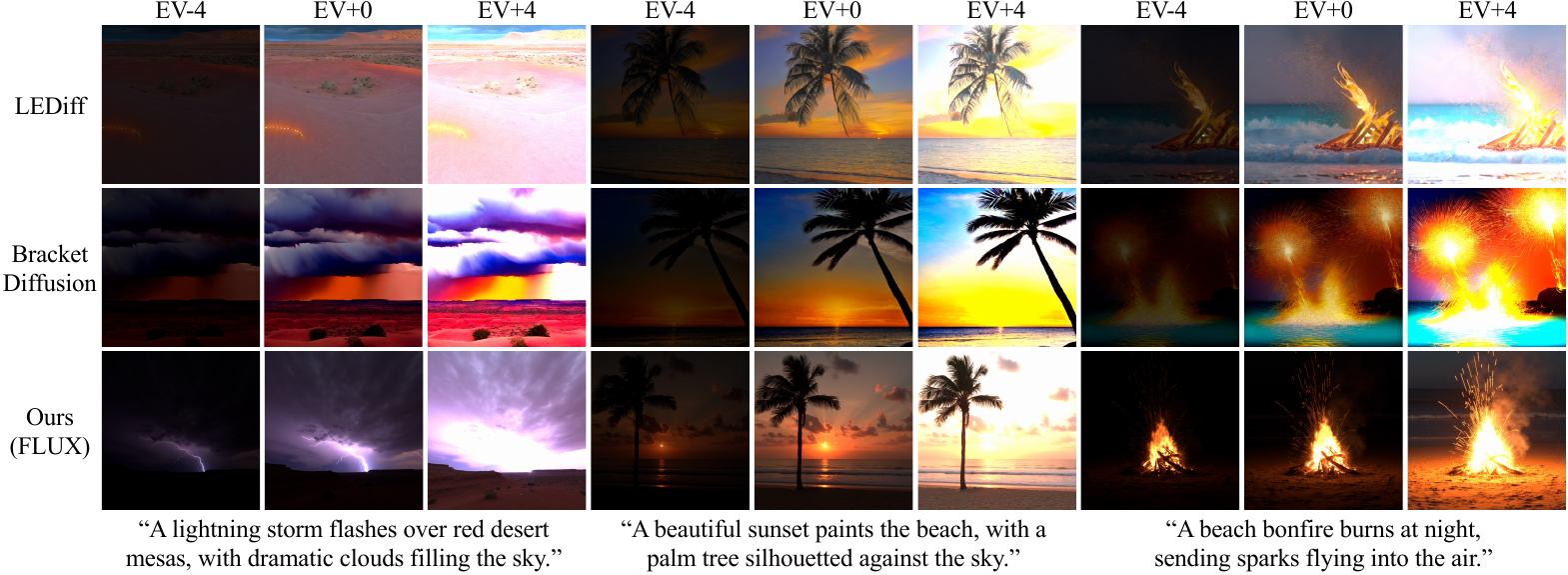}
    \caption{Visual comparison on text-to-HDR generation shown at three exposure settings (EV $-4$, EV $+0$, and EV $+4$). \sysName\ better preserves light sources at low exposure (EV $-4$) and reveals informative shadow details at high exposure (EV $+4$), demonstrating a wider effective dynamic range. In contrast, LEDiff and Bracket Diffusion often underestimate peak luminance  (dimming highlights at EV $-4$) and/or lose structure in underexposed areas (\eg, shadow collapse in the palm tree example for Bracket Diffusion).
    }
    \vspace{-1em}
\label{fig:comparison_text2hdr}
\end{figure}

\begin{figure}[t]
    \centering
    \includegraphics[width=\linewidth]{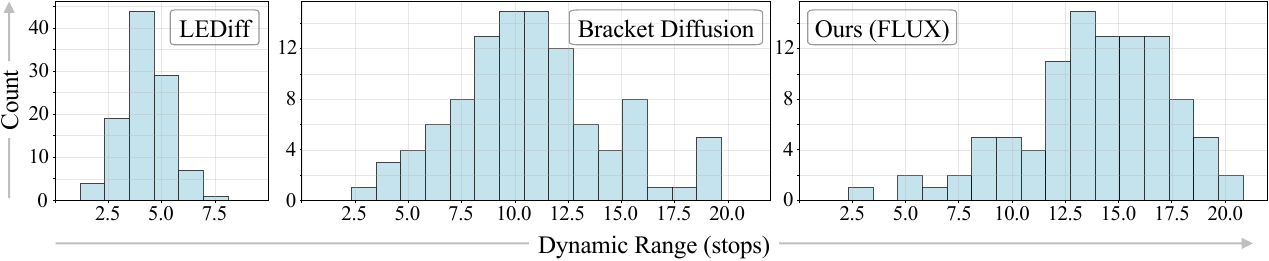}
    \caption{Distributions of effective dynamic range (in stops) over $100$ generated HDR images.
    }
\label{fig:comparison_dr}
\end{figure}

\subsection{Text-to-HDR Comparison}
\label{sec:exp_text2hdr}

We compare \sysName\ instantiated with two backbones (\texttt{FLUX.1-dev} and \texttt{SD-1.5}) against LEDiff~\cite{wang2025lediff} and Bracket Diffusion~\cite{bemana2025bracket}.
We evaluate on $100$ text prompts generated by \texttt{ChatGPT-5}, spanning diverse lighting conditions (\eg, sunlight, candles, aurora, and fireworks). To match baseline constraints, all methods generate $512{\times}512$ outputs for this comparison, though \sysName\ supports higher resolutions. The full set of prompts, additional details for the \texttt{SD-1.5} variant, and results at higher resolutions are provided in Supplemental Sections~\ref{sec_supp:hdr_results}, \ref{sec_supp:text2hdr_sd}, and \ref{sec_supp:different_resolution}.

\paragraph{Quantitative Comparison}
Table~\ref{tab:text_to_hdr} shows the quantitative results. \sysName\ achieves the best perceptual image quality and text-image alignment, with the \texttt{FLUX}-based variant performing strongest. Even when controlling for backbone (\ie, \texttt{SD-1.5}), \sysName\ remains competitive or better than competing methods while being much lighter at inference, owing to its parameter-efficient finetuning.

Beyond Q-Eval-100K scores, \sysName\ produces HDR outputs with a wide effective dynamic range ($\approx$14 stops). In contrast, LEDiff yields significantly lower dynamic range (see also Figure~\ref{fig:comparison_dr}). Bracket Diffusion reaches higher $\mathrm{DR}_\mathrm{stops}$ than LEDiff, but does so at extreme computational cost (\ie, hundreds of seconds per image) and can exhibit luminance pathologies despite high $\mathrm{DR}_\mathrm{stops}$ statistics. For example, Bracket Diffusion may collapse large regions toward near-zero values, \ie, a ``shadow clipping'' failure mode (see the palm tree in the second row of Figure~\ref{fig:comparison_text2hdr}), whereas \sysName\ tends to maintain a more natural luminance distribution for each generation.

\paragraph{Qualitative Comparison}
Figure~\ref{fig:comparison_text2hdr} visualizes results at multi-exposure settings (in the sRGB color space). At EV $-4$, \sysName\ better preserves distinct highlight emitters (\eg, lightning, bonfires, and the sun), indicating its capability of predicting sufficiently high peak luminance; LEDiff and Bracket Diffusion often appear dimmer with reduced highlight contrast.
At EV $+4$, \sysName\ reveals additional shadow structure, whereas LEDiff often washes out toward white and Bracket Diffusion tends to leave large regions nearly black due to shadow clipping.
Overall, \sysName\ better retains usable details across both highlights and shadows.

\begin{figure*}[t]
    \centering
    \includegraphics[width=\textwidth]{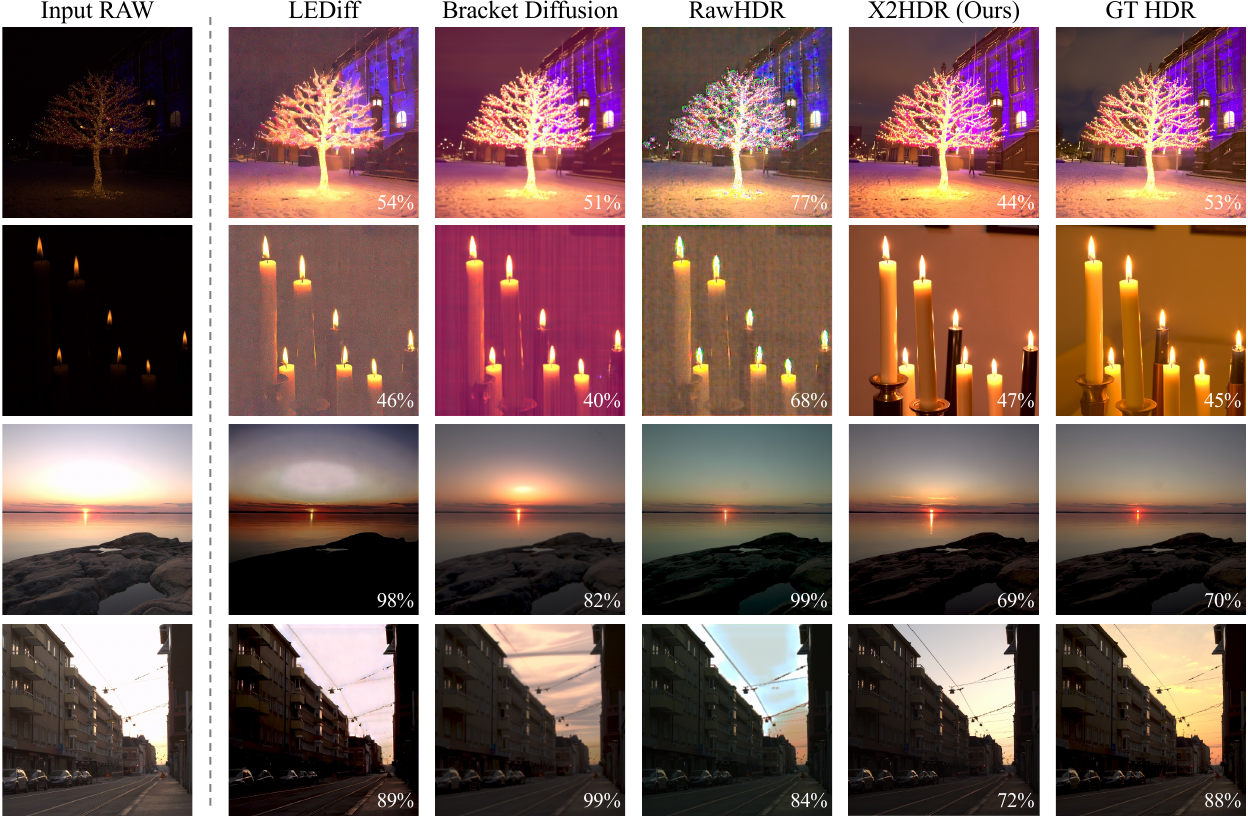}
    \caption{
    Visual comparison on RAW-to-HDR reconstruction. The competing methods exhibit similar failure modes: RawHDR is prone to noise amplification and color instability at low exposure, LEDiff often trades fidelity for excessive smoothing, and Bracket Diffusion's low conditioning resolution tends to soften edges and attenuate fine structure after upsampling. \sysName, on the other hand, better balances denoising and detail preservation, and more consistently restores plausible content in clipped regions (\eg, saturated sky) while maintaining stable color appearance. For better visual comparison, we map HDR outputs to LDR using the inverse display model of Mantiuk and Heidrich~\cite{mantiuk2009visualizing}, optimizing the respective exposures (parameterized by luminance percentiles) to best align perceived brightness relative to the selected GT HDR exposure.
    }
\label{fig:comparison_raw2hdr}
\end{figure*}

\subsection{RAW-to-HDR Comparison}
\label{sec:exp_raw2hdr}
For RAW-to-HDR, we compare \sysName\ to RawHDR~\cite{zou2023rawhdr} and additionally include LEDiff and Bracket Diffusion, adapting them to the RAW setting by a RAW-to-sRGB conversion before applying their original pipelines.
Evaluation uses  $96$ RAW images from the SI-HDR dataset~\cite{hanji2022comparison} at $512{\times}512$ (see Supplemental Section~\ref{sec_supp:raw2hdr_testset}). For methods supporting text conditioning, we use an empty prompt (\ie, setting $c_p = \varnothing$) to ensure fairness. Text-guided hallucination results are deferred to Supplemental Section~\ref{sec_supp:text_guided_hallucination}.

\begin{table}[t]
  \centering
  \footnotesize
  \caption{
  Quantitative results for RAW-to-HDR.
  }
  \label{tab:raw_to_hdr}
  \begin{tabular}{lccccc}
    \toprule
    Method
    & JOD $\uparrow$
    & $Q^\star_{\mathrm{PSNR}}\uparrow$
    & $Q^\star_{\mathrm{SSIM}}\uparrow$
    & $Q^\star_{\mathrm{LPIPS}}\downarrow$
    & $Q^\star_{\mathrm{DISTS}}\downarrow$ \\
    \midrule
    RawHDR
    & $6.58$ & $19.3$ & $0.591$ & $0.282$ & $0.133$ \\
    LEDiff
    & $6.74$ & $19.7$ & $0.570$ & $0.319$ & $0.151$ \\
    Bracket Diffusion
    & $7.08$ & $21.6$ & $0.727$ & $0.282$ & $0.148$ \\
    \sysName~(Ours)
    & $\mathbf{7.57}$ & $\mathbf{22.5}$ & $\mathbf{0.741}$ & $\mathbf{0.174}$ & $\mathbf{0.089}$ \\
    \bottomrule
  \end{tabular}
  \vspace{-0.5em}
\end{table}

\paragraph{Quantitative Comparison}
Table~\ref{tab:raw_to_hdr} reports the JOD scores by ColorVideoVDP~\cite{Mantiuk_2024_cvvdp} and exposure-optimized $Q^\star$ metrics~\cite{cao2024perceptual}. \sysName\ achieves the best performance across all reported measures, outperforming RawHDR as well as the LDR-to-HDR counterparts in this RAW setting.
These trends are consistent with the methodological differences:
RawHDR trains a small U-Net from scratch, limiting its ability to plausibly inpaint missing content, while LEDiff and Bracket Diffusion operate with weaker priors and face a larger domain gap when driven by sRGB-converted RAW inputs.

\paragraph{Qualitative Comparison}
Figure~\ref{fig:comparison_raw2hdr} shows representative results on four test RAW images.
For underexposed inputs, RawHDR is inclined to introduce noticeable chromatic noise (\eg, in the tree of the first row).  For overexposed scenes, RawHDR often suffers from color shifts (\eg, green cast) and struggles to inpaint coherent sky texture. LEDiff and Bracket Diffusion likewise fail to hallucinate plausible bright-region content. When conditioned on image inputs, Bracket Diffusion is further constrained by a lower operating resolution ($256{\times}256$), leading to blur after upsampling. In contrast, \sysName\ suppresses noise in dark regions and performs more spatially consistent inpainting in saturated areas, yielding cleaner and more realistic HDR reconstructions overall.

\begin{table*}[t]
  \centering
  \footnotesize
  \setlength{\tabcolsep}{1.9pt}
  \caption{
  Ablation results for text-to-HDR and RAW-to-HDR.
  }
  \vspace{-0.7em}
  \label{tab:ablation}

  \begin{subtable}[t]{0.47\textwidth}
    \centering
    \caption{Text-to-HDR}
    \vspace{-0.7em}
    \resizebox{\linewidth}{!}{
    \begin{tabular}{lccc}
      \toprule
      Configuration
      & \begin{tabular}[c]{@{}c@{}}Image\\Quality\end{tabular} $\uparrow$
      & \begin{tabular}[c]{@{}c@{}}Text-Image\\Alignment\end{tabular} $\uparrow$
      & \begin{tabular}[c]{@{}c@{}}Dynamic\\Range\end{tabular} $\uparrow$ \\
      \midrule
      HDR (Linear)
      & $0.578$ & $0.744$ & $5.5$ \\
      HDR (Gamma)
      & $0.629$ & $0.813$ & $10.4$ \\
      HDR (Log)
      & $0.627$ & $0.799$ & $10.5$ \\
      HDR (PQ)
      & $0.661$ & $0.822$ & $\mathbf{14.5}$ \\
      \midrule
      HDR (PU21, default)
      & $\mathbf{0.668}$ & $\mathbf{0.831}$ & $14.2$ \\
      \bottomrule
    \end{tabular}
    }
  \end{subtable}%
  \hfill
  \begin{subtable}[t]{0.51\textwidth}
    \centering
    \caption{RAW-to-HDR}
    \vspace{-0.7em}
    \resizebox{\linewidth}{!}{
    \begin{tabular}{lccccc}
      \toprule
      Configuration
      & JOD $\uparrow$
      & $Q^\star_{\mathrm{PSNR}}\uparrow$
      & $Q^\star_{\mathrm{SSIM}}\uparrow$
      & $Q^\star_{\mathrm{LPIPS}}\downarrow$
      & $Q^\star_{\mathrm{DISTS}}\downarrow$ \\
      \midrule
      HDR (Linear)
      & $5.60$ & $12.9$ & $0.306$ & $0.551$ & $0.302$ \\
      HDR (Gamma)
      & $5.99$ & $14.9$ & $0.435$ & $0.436$ & $0.222$ \\
      HDR (Log)
      & $6.12$ & $15.8$ & $0.476$ & $0.416$ & $0.208$ \\
      HDR (PQ)
      & $7.48$ & $22.3$ & $0.733$ & $0.178$ & $0.092$ \\
      \midrule
      HDR (PU21, default)
      & $\mathbf{7.57}$ & $\mathbf{22.5}$ & $\mathbf{0.741}$ & $\mathbf{0.174}$ & $\mathbf{0.089}$ \\
      \bottomrule
    \end{tabular}
    }
  \end{subtable}
  \vspace{-1em}
\end{table*}

\begin{figure*}[t]
    \centering
    \includegraphics[width=\textwidth]{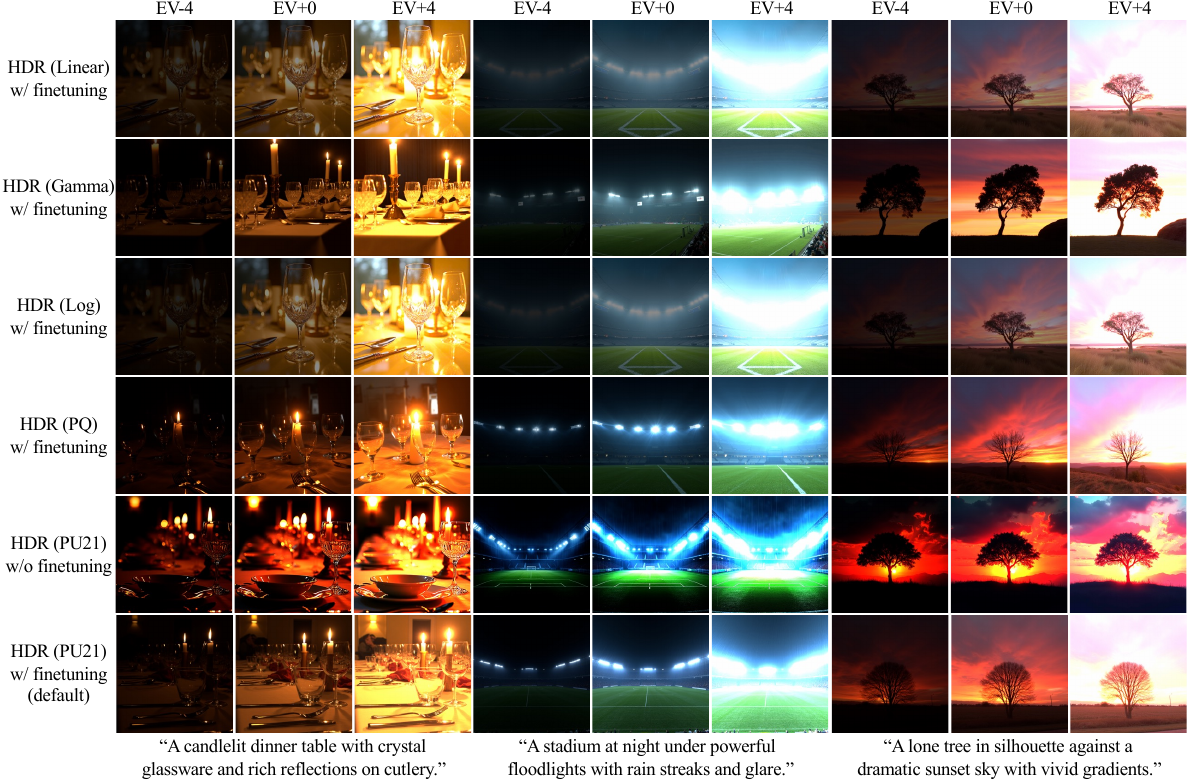}
    \caption{Text-to-HDR ablation on HDR representation and finetuning. Linear/gamma/log encodings severely compress bright emitters, most evident at EV $-4$, leading to a reduced effective dynamic range and poorer highlight/shadow recoverability. PQ closely matches PU21, preserving plausible details, whereas removing finetuning yields a ``pseudo-HDR'' appearance with exaggerated saturation/contrast and visible clipping artifacts.
    }
    \vspace{-0.5em}
\label{fig:ablation_text2hdr}
\end{figure*}

\begin{figure*}[t]
    \centering
    \includegraphics[width=\textwidth]{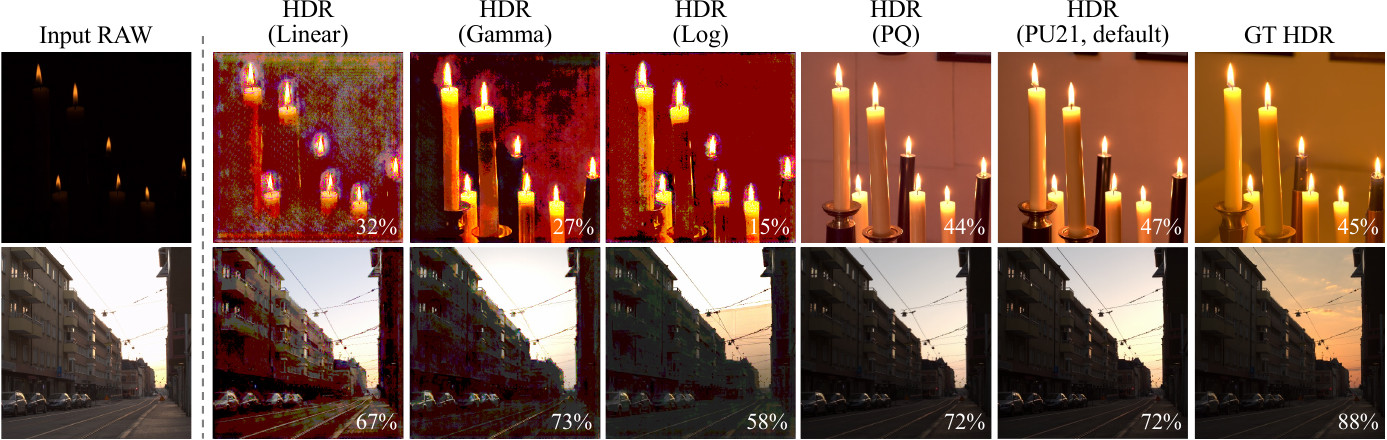}
    \caption{RAW-to-HDR ablation on HDR representation.
    Linear/gamma/log encodings introduce pronounced quantization and color distortions, whereas PQ and PU21 achieve comparable perceptual quality, with more faithful highlight recovery and reduced artifacts. In the candle scene, the background content varies between runs despite similar encoding behavior, explained by the stochasticity of the inpainting process.
    }
    \vspace{-1em}
\label{fig:ablation_raw2hdr}
\end{figure*}

\subsection{Ablation Studies}
\label{sec:exp_ablation_study}

\paragraph{Necessity for Perceptually Uniform Representation}
To isolate the role of HDR representation, we replace the default PU21 encoding with linear encoding (normalized to $[0,1]$), gamma encoding ($\gamma=2.2$), log encoding ($\mu=500$), and PQ encoding~\cite{miller2013perceptual}.
As shown in Table~\ref{tab:ablation} and Figures~\ref{fig:ablation_text2hdr} and~\ref{fig:ablation_raw2hdr}, all non-perceptually-uniform alternatives substantially degrade HDR behavior: in text-to-HDR, linear encoding yields severely limited dynamic range (\ie, $\mathrm{DR}_{\text{stops}} = 5.5$), and in RAW-to-HDR, it introduces strong artifacts (\eg, quantization), causing large drops across reconstruction metrics.
Gamma and log encodings mitigate these issues to some extent, but both still exhibit reduced dynamic range and noticeable quantization artifacts.
This arises because these representations assign an excessive portion of the available output range to highlights, resulting in a significant mismatch with sRGB statistics and lower image quality and text-image alignment scores under Q-Eval-100K.
PQ, as expected, performs comparably to PU21 on both tasks, supporting the central claim that perceptually uniform HDR representation is critical for adapting LDR-pretrained T2I models to HDR synthesis and reconstruction.

\paragraph{Necessity for Finetuning}
We also test whether HDR outputs can be obtained without any finetuning by directly decoding and rescaling pretrained T2I latents, followed by the inverse PU21 transform. This text-to-HDR configuration frequently over-stretches the tonal and chromatic range, leading to two characteristic failure modes: 1) exaggerated global contrast and saturation caused by over-expanding LDR latents (see the third row of Figure~\ref{fig:ablation_text2hdr}) and 2) widespread clipping that removes recoverable details (\eg, the grassland remains dark even at EV $+4$).
We direct the readers to more visual examples in the HTML supplementary and the discussion in Supplemental Section~\ref{sec_supp:q_eval}.

\section{Conclusion}
We have presented \sysName, a simple and effective adaptation method that enables pretrained T2I diffusion models to operate in HDR for both text-to-HDR generation and RAW-to-HDR reconstruction. The key idea is to mitigate the distribution gap between linear-light HDR/RAW data and LDR-pretrained models by mapping HDR values into a perceptually uniform space. This simple preprocessing allows an off-the-shelf LDR-pretrained VAE to represent HDR content faithfully, so HDR capability can be acquired by parameter-efficient LoRA finetuning of the denoiser while keeping the rest of the computational structure unchanged. The resulting \sysName\ is unified and deployment-friendly, and it yields reliable HDR generation and reconstruction results with improved perceptual fidelity and usable dynamic range.

\section*{Acknowledgments}
This work was supported in part by the National Natural Science Foundation of China and the Research Grants Council of Hong Kong under the Collaborative Research Scheme (Project No. CRS\_HKUST605/25), and by the Innovation and Technology Fund of the Hong Kong Innovation and Technology Commission (Project Nos. 9440379 and 9440390).

{
\small
\bibliographystyle{plain}
\bibliography{ref}
}

\appendix

\section{Overview}
This appendix provides supplementary results and implementation details that support and extend the findings of the main paper.
Specifically, it includes:
\begin{itemize}
    \item Prior HDR generation and reconstruction methods (Section~\ref{sec_supp:related_work});
    \item Complete HDR results for text-to-HDR generation and RAW-to-HDR reconstruction, including comparisons, ablations, and perceptual experiments, as well as HDR outputs at multiple resolutions and for synthetic stimuli (Sections~\ref{sec_supp:hdr_results}, \ref{sec_supp:text2hdr_sd}, \ref{sec_supp:raw2hdr_synthetic}, and~\ref{sec_supp:different_resolution});
    \item Detailed analysis of PU21 perceptually uniform encoding (Section~\ref{sec_supp:pu21_explanation});
    \item HDR encoding function visualization (Section~\ref{sec_supp:encoding_functions});
    \item Dataset and reproducibility details for training and evaluation (Sections~\ref{sec_supp:training_dataset} and~\ref{sec_supp:implementation_details});
    \item Extended VAE analysis, including failure characterization, VAE finetuning, and latent distribution visualization (Section~\ref{sec_supp:more_vae_experiments});
    \item LDR-to-HDR reconstruction experiments (Section~\ref{sec_supp:ldr2hdr});
    \item Perceptual study details (Section~\ref{sec_supp:perceptual_study});
    \item Applications to text-guided hallucination (Section~\ref{sec_supp:text_guided_hallucination});
    \item Limitations (Section~\ref{sec_supp:limitations}) and future research directions (Section~\ref{sec_supp:future_work}).
\end{itemize}

\section{Related Work}
\label{sec_supp:related_work}
Our work connects three lines of research: T2I diffusion models in LDR, efforts to extend such models to  HDR, and HDR reconstruction from a single image or multi-exposure brackets.

\subsection{T2I Diffusion Models in LDR}
Diffusion models have become a dominant paradigm for text-guided image synthesis, producing high-quality results across diverse scene types, semantic compositions, and visual styles~\cite{chen2023pixart, rombach2022high, saharia2022photorealistic, nichol2021glide}. Latent diffusion models~\cite{rombach2022high} improve scalability by learning the generative process in a compressed latent space: a VAE maps high-dimensional images into low-dimensional latents, and a denoiser learns to invert progressive corruption in this space. More recent work integrates Transformer backbones~\cite{vaswani2017attention} into diffusion processes, giving rise to diffusion Transformers (DiT)~\cite{peebles2023scalable}. 
Building on these developments, \texttt{FLUX}~\cite{blackforestlabs_flux} combines Transformer-based architectures with flow matching objectives~\cite{lipman2022flow, albergo2022building, liu2022flow}, achieving state-of-the-art T2I generation performance. 
Here we mainly adopt \texttt{FLUX.1-dev} as the backbone and investigate how to extend it to HDR tasks with minimal changes.

\subsection{HDR Image Generation}
Compared with LDR generation, text-guided HDR image synthesis remains relatively underexplored. Two closely related methods are LEDiff~\cite{wang2025lediff} and Bracket Diffusion~\cite{bemana2025bracket}. Both emulate the classical HDR workflow by generating multi-exposure ``brackets'' and subsequently merging them into a single HDR output~\cite{debevec2023recovering}. LEDiff finetunes specialized denoisers for highlight hallucination and shadow recovery, and adapts the VAE decoder to produce linear RGB values. Bracket Diffusion instead jointly denoises over multiple exposures, incorporating constraints that preserve exposure diversity while enforcing semantic consistency across the exposure stack. While effective, these bracket-and-merge designs add architectural and algorithmic complexity, increase inference-time and memory requirements, and can introduce unwanted fusion artifacts---factors that hinder portability to newer, more memory-intensive T2I backbones.
In contrast, \sysName\ adapts pretrained T2I models by operating directly in a perceptually uniform space, avoiding explicit bracket generation and fusion while remaining compatible with modern architectures.

\subsection{HDR Image Reconstruction}
Most HDR imaging techniques rely on multi-exposure bracketing: capturing several LDR exposures and merging them into HDR radiance maps~\cite{debevec2023recovering}. For dynamic scenes, however, camera/scene motion causes misalignment and ghosting, motivating an extensive body of work on motion-aware HDR reconstruction~\cite{sen2012robust}. Recent methods integrate stronger alignment and deghosting modules~\cite{kalantari2017deep, yan2019attention,yan2023smae, chen2025ultrafusion, kong2024safnet,ye2021progressive} with powerful backbones such as U-Nets~\cite{wu2018deep} and Transformers~\cite{chen2023improving,liu2022ghost,song2022selective,tel2023alignment}. Nevertheless, robust reconstruction under large motion and severe saturation remains challenging.

A complementary line of research considers single-image LDR-to-HDR reconstruction (also referred to as inverse tone-mapping). Classical inverse methods expand dynamic range via heuristic linearization and optional detail hallucination~\cite{akyuz2007hdr,banterle2008expanding,banterle2009psychophysical,didyk2008enhancement,masia2009evaluation,masia2017dynamic}. Learning-based methods either approximately invert the ISP end-to-end~\cite{eilertsen2017hdr,liu2020single,marnerides2018expandnet,santos2020single,yu2021luminance, dille2024intrinsic}, or synthesize multi-exposures from a single image and then apply conventional HDR merging~\cite{endo2017deep,lee2018deep,zhang2023revisiting,wang2025lediff,bemana2025bracket}. However, limited training data can reduce generalization, and bracket-synthesis approaches tend to introduce inter-exposure inconsistencies, especially under extreme dark and bright regions.

In this paper, we depart from the prevailing LDR-to-HDR paradigm and study HDR reconstruction from a single RAW capture, a setting that has received comparatively little attention. 
Zou~\etal~\cite{zou2023rawhdr} explored a related setting with a rather elaborate workflow. In contrast, we propose a simple approach that leverages powerful generative priors to denoise and inpaint effectively during HDR reconstruction, yielding superior performance.

\section{Complete HDR Results}
\label{sec_supp:hdr_results}
Because HDR content is difficult to reproduce faithfully in a static PDF, we provide extended results on the project website {\color{blue}\url{https://X2HDR.github.io/}}, including 1) $100$ text-to-HDR images, 2) $96$ RAW-to-HDR reconstructions, 3) ablation results under linear encoding, PQ encoding, and without finetuning for both text-to-HDR and RAW-to-HDR tasks, and 4)  visual stimuli selected in the perceptual study with their corresponding JOD scores.

To ensure comparable perceived brightness, we normalize all HDR images by mapping the median luminance to $0.5~\mathrm{cd/m^2}$.
For efficient browser-based viewing, EXR files are converted to a compact JPEG gain-map representation using \texttt{gainmap-js}\footnote{\url{https://github.com/MONOGRID/gainmap-js}}. Note that a correct HDR viewing experience requires an HDR-capable display; on LDR monitors, appearance may deviate from true HDR content, and perceived quality depends on the specific HDR hardware.

\section{PU21 Perceptually Uniform Encoding}
\label{sec_supp:pu21_explanation}

PU21~\cite{Mantiuk2021} is a scalar encoding that maps absolute HDR luminance values from $0.005$ to $10{,}000$\cdms{} into an approximately perceptually uniform domain. In this domain, Euclidean distances correspond more closely to visible differences, allowing metrics such as PSNR and SSIM to be applied meaningfully to HDR images. The encoding is derived by numerically integrating contrast sensitivity, defined as the inverse of the detection threshold, over log-luminance. An analytic, invertible curve is then fitted and scaled to map the luminance range to $[0,1]$.

Whereas PU08~\cite{aydin2008extending} and PQ~\cite{miller2013perceptual} are derived from peak contrast sensitivity, PU21 models the detection of banding artifacts caused by fixed-bit-depth quantization. It also accounts for the reduced visibility of differences in dark image regions due to disability glare---the scattering of light within the optics of the eye.

\begin{figure}[t]
    \centering
    \includegraphics[width=0.7\linewidth]{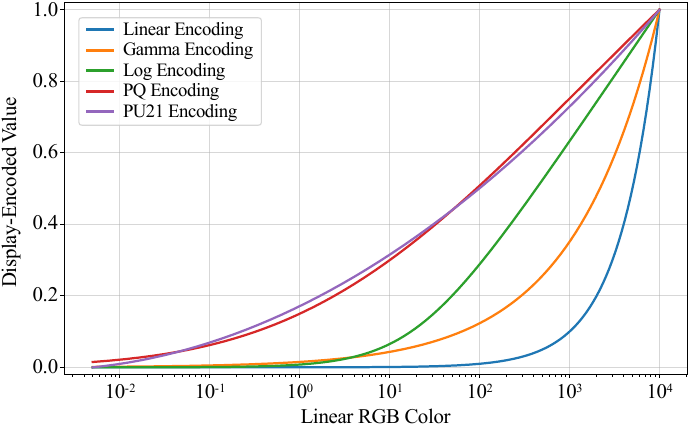}
    \caption{
    Encoding functions (linear, gamma, log, PQ, and PU21) mapping BT.2100-range linear luminance to display-encoded values (with log-scaled x-axis).
    }
    \label{fig:encoding_function}
\end{figure}

\begin{figure}[t]
    \centering
    \includegraphics[width=0.8\linewidth]{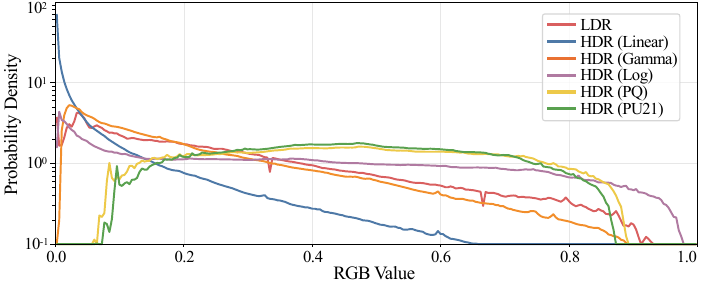}
    \caption{
    RGB value distributions under different encodings.
    }
    \label{fig:vae_analysis_supp}
\end{figure}

\section{Encoding Function Visualization}
\label{sec_supp:encoding_functions}

Figure~\ref{fig:encoding_function} compares five encodings used to map linear RGB luminance to display-encoded values: linear encoding, gamma encoding ($\gamma=2.2$), log encoding ($\mu=500$), PQ~\cite{miller2013perceptual}, and PU21~\cite{Mantiuk2021}.
Linear mapping assigns a disproportionately large range to highlights, heavily compressing low-luminance values.
Gamma and log encodings partially rebalance this allocation, but they still do not achieve the same degree of perceptual uniformity as PQ and PU21.
In contrast, PQ and PU21 nonlinearly compress extreme highlights while allocating more resolution to shadows and midtones, spreading low-luminance differences in the encoded domain.
Although PQ and PU21 differ in functional form, both reshape HDR statistics toward distributions that better match LDR-pretrained models.
We further visualize the RGB value distributions in Figure~\ref{fig:vae_analysis_supp} using the histogram procedure described in Section~\ref{sec:method_vae_recon}.
Gamma and log encodings produce distributions closer to that of LDR data than PQ and PU21 do.
However, the ablation study in Section~\ref{sec:exp_ablation_study} shows that they underperform perceptually uniform encodings after finetuning, suggesting that LDR-like statistics alone are insufficient for HDR adaptation. One possible explanation lies in the inverse transformation applied after generation: the decoded image must be mapped back to linear RGB values, and this mapping can amplify errors introduced in the encoded domain.
Although generic concave mappings such as gamma and log improve histogram alignment, they do not allocate encoded values according to perceptual sensitivity across the HDR luminance range.
Consequently, small errors in the encoded domain can produce visible errors after inversion.

\section{More Experimental Details and Results}
\subsection{Training Dataset}
\label{sec_supp:training_dataset}

\paragraph{Text-to-HDR}
We build the training set ($3{,}278$ HDR images) primarily from publicly available HDR datasets used in prior work~\cite{polyhaven, bolduc2023beyond, fairchild2007hdr, gardner2017learning, kalantari2017deep, liu2020single, panetta2021tmo, tel2023alignment}.
From each HDR image, we randomly extract $512{\times}512$ and $1,024{\times}1,024$ crops, retaining only those with an effective dynamic range of $\mathrm{DR}_{\text{stops}} \ge 5$.
For HDR panoramas, random perspective projections are applied before cropping. To better exploit high-resolution sources, we use an adaptive sampling scheme that draws proportionally more crops from larger images, producing $32,200$ patches at $512\times 512$ and $6,714$ patches at $1,024\times 1,024$. Text prompts are obtained by tone-mapping HDR crops~\cite{drago2003adaptive}  and captioning them with a VLM (\ie, \texttt{Gemini-2.0-Flash}) using the following instruction:
\begin{lstlisting}
You are a professional image-captioning assistant. Generate objective, accurate, and detailed captions based on the provided image. Produce two outputs: (i) a short caption (1--3 sentences) that summarizes the main content, and (ii) a long caption (one paragraph) that describes all salient details. Use precise, descriptive language; remain factual and avoid subjective interpretation. Output only the captions, formatted exactly as: '###Short:' and '###Long:'.
\end{lstlisting}
During training, we randomly select either the short or long caption at each iteration.

\paragraph{RAW-to-HDR}
The RAW-to-HDR training data are drawn from prior HDR datasets~\cite{fairchild2007hdr, zou2023rawhdr}, comprising $517$ scenes with multiple RAW exposures.
For each scene, bracketed RAW images are merged into a reference HDR image using HDRUtils\footnote{\url{https://github.com/gfxdisp/HDRUtils}}, with exposure estimation and alignment~\cite{hanji2020noise, hanji2023exposures}. Paired (RAW, HDR) samples are formed by cropping spatially co-located regions. We extract $10$ crops per RAW exposure, resulting in a total of $22{,}160$ pairs at $512{\times}512$. HDR targets are tone-mapped and captioned with the same VLM.

\subsection{RAW-to-HDR Test Dataset}
\label{sec_supp:raw2hdr_testset}

Instead of splitting the collected training data, evaluation is performed on an independent SI-HDR dataset~\cite{hanji2022comparison}, consisting of $183$ scenes with up to $7$ RAW exposures and merged HDR references. It provides a clear domain gap and is well-suited for assessing the generalizability of RAW-to-HDR reconstruction methods.
We randomly sample $96$ scenes; for each scene, we discard the darkest and brightest exposures and then randomly choose one RAW image from the remaining. Each selected RAW image is cropped and resized to ensure pixel-alignment with the HDR reference at $1,888{\times}1,280$, and then further cropped and downsampled to $512\times 512$ for evaluation.

\subsection{Implementation Details}
\label{sec_supp:implementation_details}

We use \texttt{FLUX.1-dev} as the default T2I backbone.
For text-to-HDR, we set the LoRA rank to $32$, with a learning rate of $1\times10^{-4}$ and a dedicated LoRA trigger token, \texttt{[PU21]}.
The per-GPU batch size is $8$ for $512{\times}512$ and $2$ for $1,024{\times}1,024$, with $4$ gradient accumulation steps.
Training converges after approximately $3{,}000$ steps on $4$ GPUs.
For RAW-to-HDR, we adopt EasyControl~\cite{zhang2025easycontrol} for conditional generation, which enhances the DiT backbone by introducing attention masking, position-encoding cloning, and conditional LoRA injection.
We set the LoRA rank to $128$ with learning rate $1\times10^{-4}$.
During training, we randomly drop text prompts (\ie, setting $c_p=\varnothing$) with  $50\%$ probability to implement RAW-only reconstruction. Training runs for $8$ epochs on $512{\times}512$ images with batch size $2$ on $4$ GPUs.
For both tasks, the LoRA scaling factor $\alpha$ is set equal to the LoRA rank.

\subsection{Additional VAE Analysis}
\label{sec_supp:more_vae_experiments}

By default, \sysName\ freezes the pretrained VAE and adapts only the denoiser of the T2I model. We further analyze when the frozen VAE fails on PU21-encoded HDR data and whether HDR-specific VAE finetuning provides additional benefits.

\begin{figure}[t]
    \centering
    \includegraphics[width=0.7\linewidth]{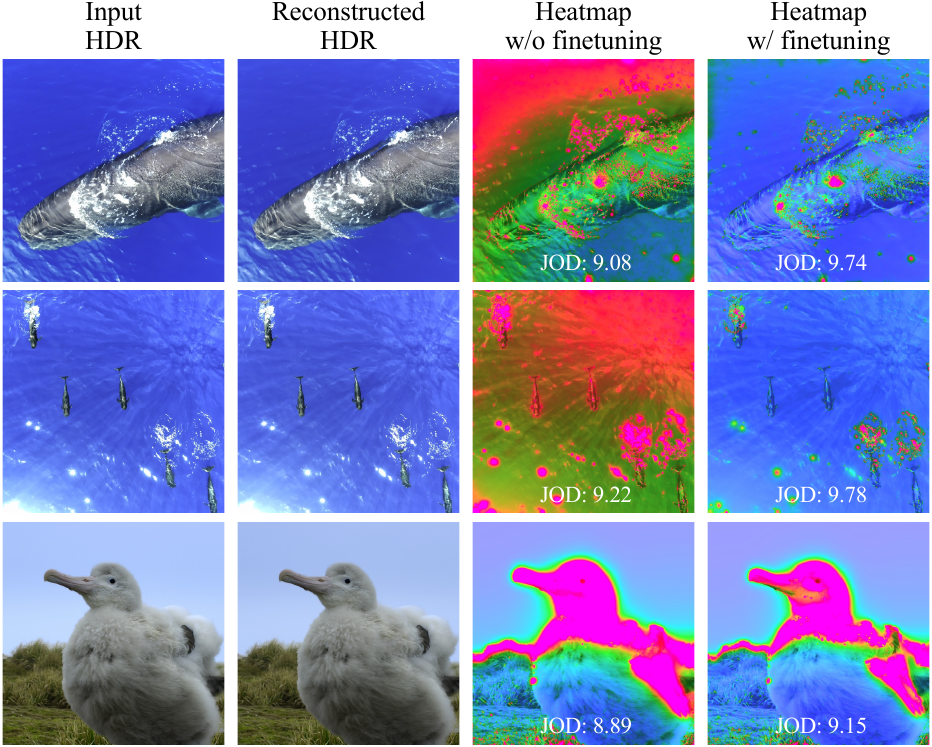}
    \caption{
    Representative VAE reconstruction failures on PU21-encoded HDR images, highlighted by ColorVideoVDP error maps.
    }
    \label{fig:vae_recon_fail}
\end{figure}

\paragraph{Failure characterization}
While PU21 generally induces high-fidelity reconstruction, the pretrained VAE occasionally under-reconstructs highlight regions (see Figure~\ref{fig:vae_recon_fail}).

\paragraph{VAE finetuning}
To improve highlight reconstruction, we finetune the pretrained \texttt{FLUX.1-dev} VAE on PU21-encoded HDR data.
Dataset preparation largely follows the text-to-HDR training pipeline (see Section~\ref{sec_supp:training_dataset}), but omits VLM captioning and fixes crop resolution to $768{\times}768$.
This yields a total of $60{,}399$ HDR images. Training uses learning rate $1\times10^{-5}$, batch size $1$, $8$ gradient accumulation steps, $2$ epochs, and $4$ GPUs. Several reconstruction losses are tested, including the
 mean absolute error (MAE), locally adaptive DISTS (ADISTS) metric~\cite{ding2021locally}, a differentiable histogram matching loss~\cite{mustafa2022comparative}, and their variants combined with an inverse display model~\cite{mantiuk2009visualizing}. MAE is found sufficient.

After finetuning, average JOD on $512$ test HDR images rises from $9.44$ to $9.72$, with visible improvements in some highlight cases (see the last column of Figure~\ref{fig:vae_recon_fail}). However, gains are not consistent across content, and the finetuned VAE remains below the LDR reconstruction counterpart. The remaining gap may be due to limited HDR data scale and diversity (with many crops originating from relatively few scenes).

\begin{figure}[t]
    \centering
    \includegraphics[width=0.8\linewidth]{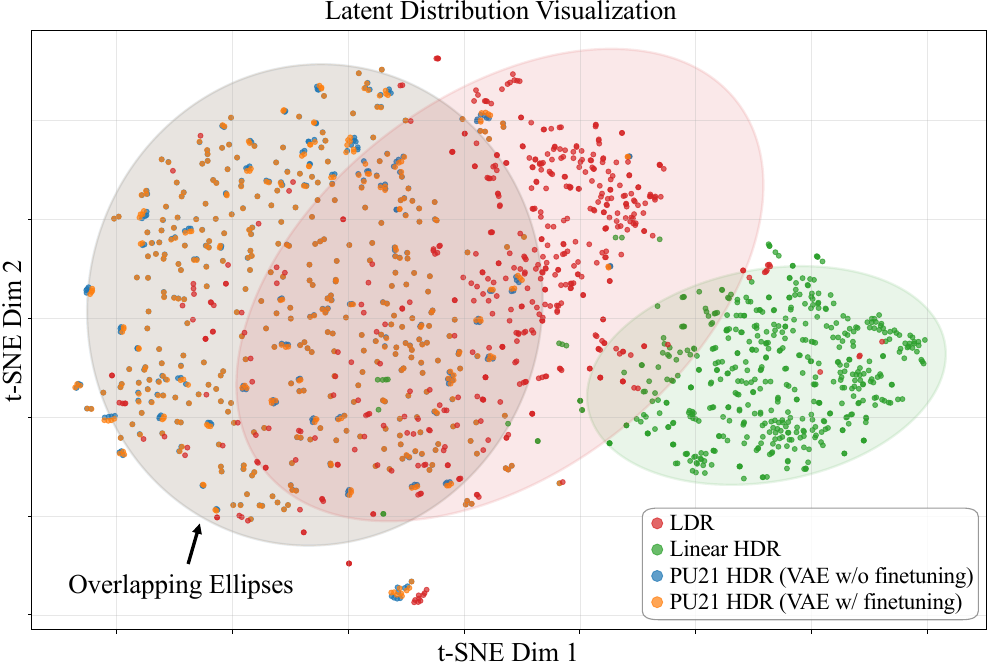}
    \caption{t-SNE visualization of VAE latent distributions under different encodings. PU21 substantially reduces the LDR-HDR mismatch, while VAE finetuning produces only marginal changes.
    }
    \label{fig:latent_visualization}
\end{figure}

\paragraph{Latent distribution visualization} Using $512$ (LDR, HDR) pairs, we compare the VAE latent distributions for 1) LDR inputs, 2) linear HDR inputs, 3) PU21-encoded HDR inputs with the pretrained VAE, and 4) PU21-encoded HDR inputs with the finetuned VAE. A t-SNE projection with $2\sigma$ confidence ellipses in Figure~\ref{fig:latent_visualization} shows that PU21 significantly reduces the LDR-HDR mismatch relative to linear HDR, while VAE finetuning produces only marginal latent changes.

\begin{table}[t]
  \centering
  \setlength{\tabcolsep}{3pt}
  \footnotesize
  \caption{Effect of VAE finetuning on RAW-to-HDR reconstruction.
  }
  \label{tab:raw_to_hdr_finetuned_vae}
  \begin{tabular}{lccccc}
    \toprule
    Method
    & JOD $\uparrow$
    & $Q^\star_{\mathrm{PSNR}}\uparrow$
    & $Q^\star_{\mathrm{SSIM}}\uparrow$
    & $Q^\star_{\mathrm{LPIPS}}\downarrow$
    & $Q^\star_{\mathrm{DISTS}}\downarrow$ \\
    \midrule
    VAE w/ finetuning
    & $7.34$ & $22.3$ & $\mathbf{0.759}$ & $0.197$ & $0.111$ \\
    VAE w/o finetuning (default)
    & $\mathbf{7.57}$ & $\mathbf{22.5}$ & $0.741$ & $\mathbf{0.174}$ & $\mathbf{0.089}$ \\
    \bottomrule
  \end{tabular}
\end{table}

\paragraph{Effect of VAE finetuning on RAW-to-HDR} Training a RAW-to-HDR model with the finetuned VAE (all else unchanged) does not yield consistent improvements. As shown in Table~\ref{tab:raw_to_hdr_finetuned_vae}, aside from a modest gain in $Q^\star_{\mathrm{SSIM}}$, other measures slightly degrade, plausibly due to mild latent distribution shift that affects denoising and inpainting during reverse diffusion.

\paragraph{Summary} Perceptually uniform encoding (\eg, PU21) is the dominant factor facilitating alignment between LDR and HDR latent representations. Under the current data scale, VAE finetuning adds limited benefits, though it may be more valuable with larger and more diverse HDR datasets.

\begin{figure}[t]
    \centering
    \includegraphics[width=0.8\linewidth]{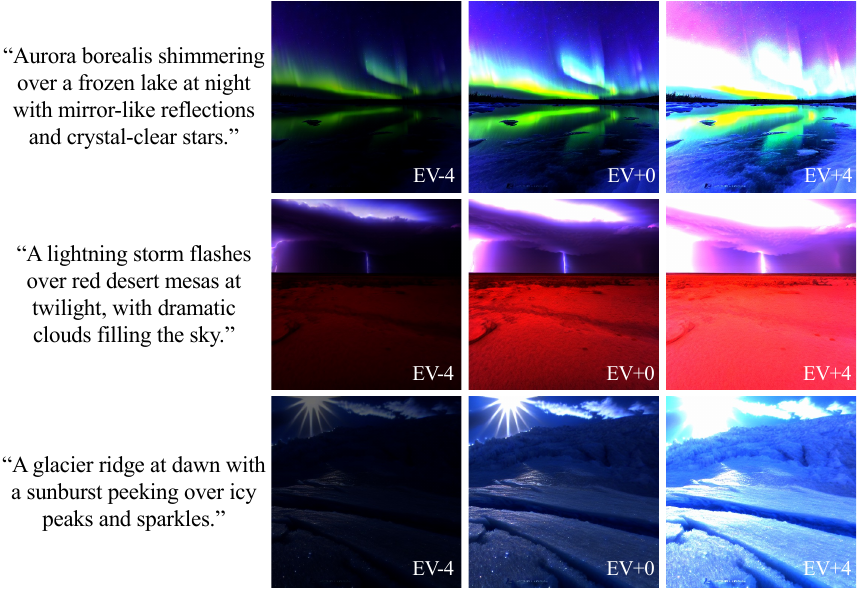}
    \caption{
    Sample HDR images generated by \sysName\ with \texttt{SD-1.5} at different exposure levels.
    }
\label{fig:text2hdr_sd}
\end{figure}

\begin{figure}[t]
    \centering
    \includegraphics[width=0.75\linewidth]{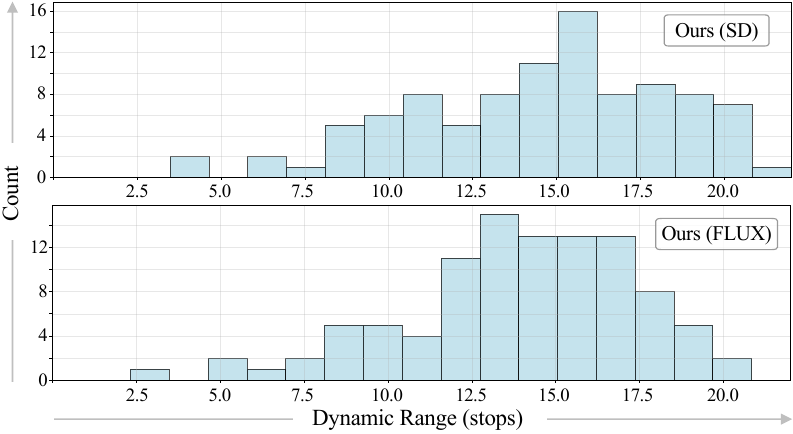}
    \caption{
    Distributions of effective dynamic range (in stops) over $100$ generated HDR images.
    Both the \texttt{SD}-based and \texttt{FLUX}-based variants produce images with wide dynamic range.
    }
\label{fig:comparison_dr_sd}
\end{figure}

\subsection{Text-to-HDR with \texttt{SD-1.5}}
\label{sec_supp:text2hdr_sd}
To verify backbone generality (and support fair comparison with LEDiff and Bracket Diffusion), \sysName\ is also instantiated with 
 \texttt{SD-1.5}, using the same PU21 encoding and LoRA strategy. The  \texttt{SD-1.5} setup uses LoRA rank $8$, learning rate $1\times10^{-5}$, and batch size $32$. Training is limited to $1{,}000$ steps to avoid overfitting, and the trigger token \texttt{[PU21]} is prepended.

Additional visual results in Figure~\ref{fig:text2hdr_sd} show that highlights remain well separated at EV $-4$ and shadow details emerge at EV $+4$. Figure~\ref{fig:comparison_dr_sd} further indicates that the \texttt{SD}-based variant achieves dynamic range comparable to the \texttt{FLUX}-based model, supporting the reliability of PU21-based adaptation across backbones.

\begin{figure*}[t]
    \centering
    \includegraphics[width=\textwidth]{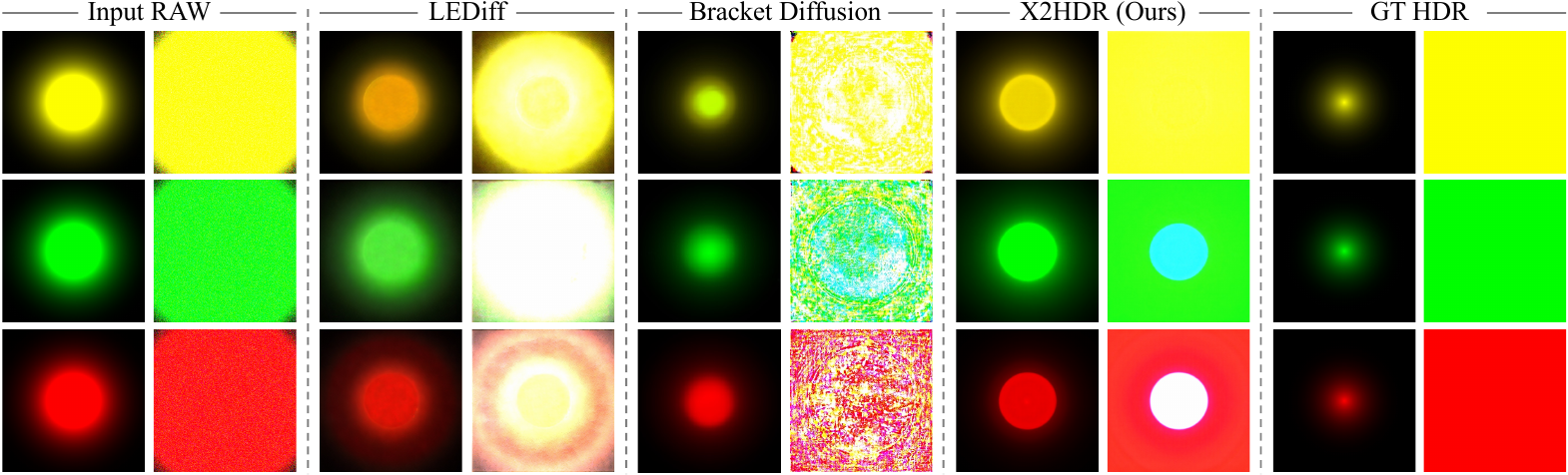}
    \caption{
    RAW-to-HDR reconstruction on synthetic data with controlled logarithmic luminance gradients.
    Rows correspond to different chromatic gradients (yellow, green, and red), and each method is visualized at two exposure levels. 
    \sysName\ strongly suppresses sensor noise, while LEDiff and Bracket Diffusion introduce structured artifacts. All methods remain challenged in recovering perfectly smooth gradients within severely clipped regions, indicating limited generalization to these synthetic patterns.
    }
    \label{fig:raw2hdr_synth}
\end{figure*}

\begin{figure*}[t]
    \centering
    \includegraphics[width=\textwidth]{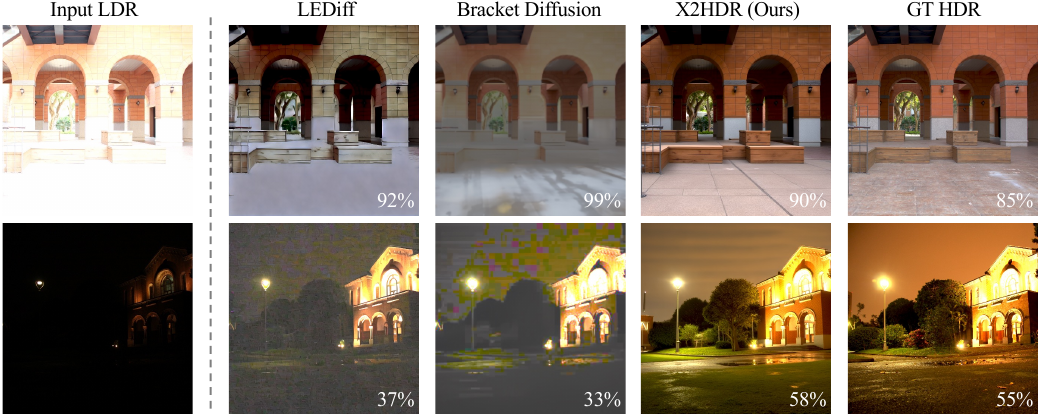}
    \caption{
    Visual comparison of LDR-to-HDR reconstruction. For the overexposed input (top), \sysName\ recovers highlight structure more plausibly, while LEDiff and Bracket Diffusion produce visible distortions. For the underexposed input (bottom), \sysName\ reveals more faithful shadow details with fewer artifacts.
    }
    \label{fig:ldr_to_hdr}
\end{figure*}

\begin{table}[t]
  \centering
  \footnotesize
  \caption{
  Quantitative results for LDR-to-HDR.
  }
  \label{tab:ldr_to_hdr}
  \begin{tabular}{lccccc}
    \toprule
    Method
    & JOD $\uparrow$
    & $Q^\star_{\mathrm{PSNR}}\uparrow$
    & $Q^\star_{\mathrm{SSIM}}\uparrow$
    & $Q^\star_{\mathrm{LPIPS}}\downarrow$
    & $Q^\star_{\mathrm{DISTS}}\downarrow$ \\
    \midrule
    LEDiff
    & $5.74$ & $17.7$ & $0.569$ & $0.252$ & $0.130$ \\
    Bracket Diffusion
    & $5.86$ & $18.4$ & $0.641$ & $0.272$ & $0.160$ \\
    \sysName~(Ours)
    & $\mathbf{6.47}$ & $\mathbf{19.3}$ & $\mathbf{0.680}$ & $\mathbf{0.187}$ & $\mathbf{0.101}$ \\
    \bottomrule
  \end{tabular}
\end{table}

\subsection{RAW-to-HDR on Synthetic Data}
\label{sec_supp:raw2hdr_synthetic}
To probe RAW-to-HDR behavior under controlled degradation, we construct a synthetic dataset using $512{\times}512$ HDR images with radial luminance gradients that decrease logarithmically from $4{,}000$\cdms{} at the center to $0.005$\cdms{} at the corners, instantiated across multiple chromatic channels (\eg, yellow and red). Synthetic RAW inputs are simulated by applying a virtual exposure that clips values above $100$\cdms{} and adding Poisson-Gaussian noise calibrated to Sony A7R~III.
 
Reconstructions in Figure~\ref{fig:raw2hdr_synth} indicate that \sysName\ effectively suppresses sensor noise, while competing techniques introduce structured artifacts. However, all methods struggle to recover ideal smooth gradients in heavily clipped regions, reflecting limited generalization to synthetic patterns absent from training.
RawHDR~\cite{zou2023rawhdr} is excluded because it requires camera-specific metadata (\eg, Bayer pattern and camera-to-RGB transformations) not defined in this synthetic setup.

\subsection{LDR-to-HDR Image Reconstruction}
\label{sec_supp:ldr2hdr}
We adapt \sysName\ to single-image LDR-to-HDR reconstruction by conditioning the model directly on LDR inputs. We train it on $9{,}786$ paired $512{\times}512$ LDR and HDR images from SingleHDR~\cite{liu2020single}, with VLM-generated captions for the HDR targets. The training setup follows that of RAW-to-HDR reconstruction, except that we reduce the number of training steps to $6{,}000$ for the smaller dataset.

For evaluation, we randomly sample $50$ images from the SingleHDR test set and compare \sysName\ with LEDiff and Bracket Diffusion. The sampled images present challenging overexposed or underexposed regions. Visual results for all $50$ test images are provided on the supplementary website.
We focus on generative baselines rather than conducting an exhaustive comparison with earlier LDR-to-HDR methods, as prior work has established the advantages of generative approaches over conventional methods trained from scratch. As shown in Table~\ref{tab:ldr_to_hdr}, \sysName\ achieves the best performance across all metrics among the evaluated generative methods. The visual comparisons in Figure~\ref{fig:ldr_to_hdr} further show that \sysName\ more faithfully reconstructs highlight structures, recovers low-light details, and produces fewer artifacts.

\begin{figure}[t]
    \centering
    \begin{minipage}[c]{0.6\textwidth}
        \centering
        \includegraphics[width=\textwidth]{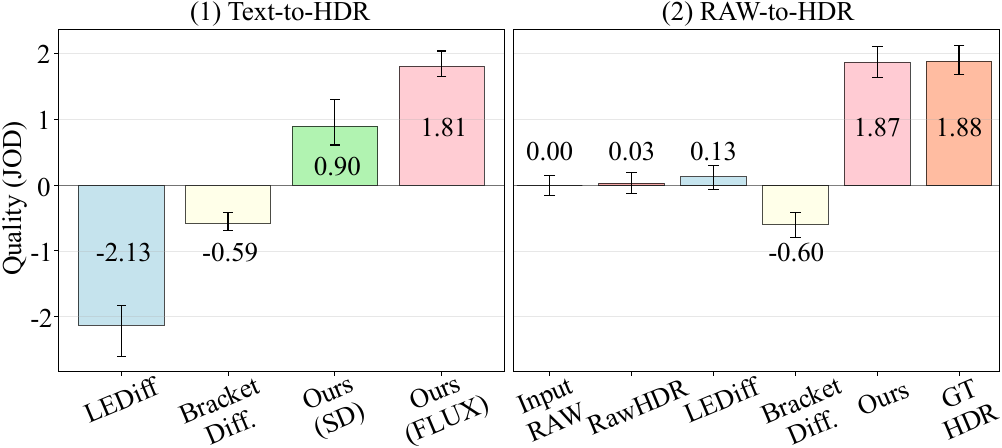}
    \end{minipage}%
    \hfill
    \begin{minipage}[c]{0.37\textwidth}
        \caption{
        Quantitative results of the perceptual study, reported in the unit of JOD (higher is better). 
        Bars show mean JOD across stimuli, and error bars denote $95\%$ confidence intervals.
        }
        \label{fig:perceptual_study}
    \end{minipage}
\end{figure}

\begin{figure*}[t]
    \centering
    \includegraphics[width=\textwidth]{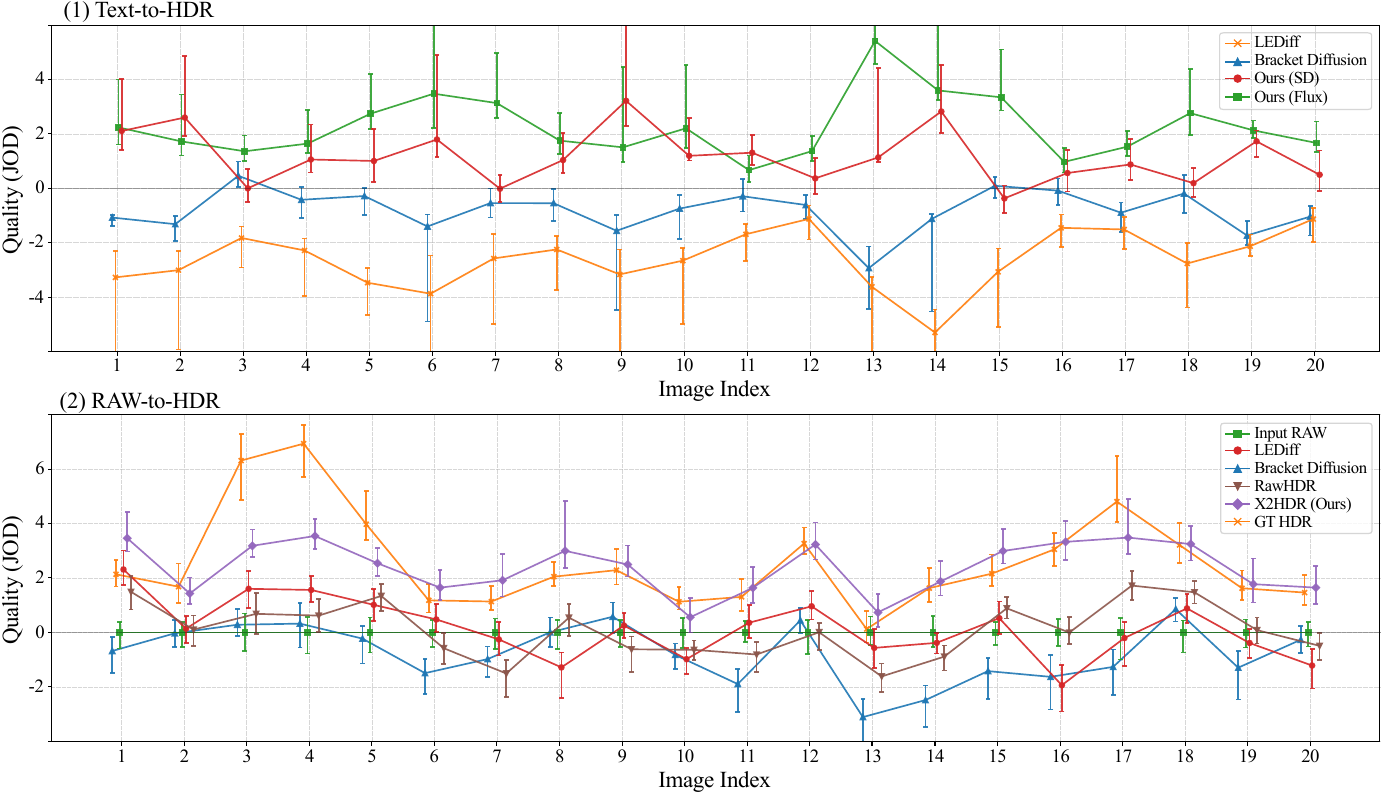}
    \caption{
    Per-image JOD scores from our perceptual study.
    The top and bottom panels report text-to-HDR and RAW-to-HDR results, respectively.
    Error bars denote $95\%$ confidence intervals via bootstrapping.
    A small horizontal jitter is applied to separate methods for improved visibility.
    }
    \label{fig:perceptual_study_supp}
\end{figure*}

\subsection{Perceptual Study}
\label{sec_supp:perceptual_study}

To complement the objective evaluations in Section~\ref{sec:exp}, we conducted a controlled perceptual study on an HDR display to verify the perceptual gains obtained by \sysName\ for both text-to-HDR generation and RAW-to-HDR reconstruction.

\paragraph{Stimuli}
For text-to-HDR, each trial compared a pair of HDR images generated from the same text prompt by two of four methods: LEDiff, Bracket Diffusion, and \sysName\ instantiated with either \texttt{FLUX.1-dev} or \texttt{SD-1.5}. We randomly selected $20$ prompts from the full prompt set.
For RAW-to-HDR, each trial compared a pair of images drawn from two of six conditions: the input RAW capture, the GT HDR reference, and outputs from LEDiff, Bracket Diffusion, RawHDR, and \sysName. We randomly sampled $20$ test images for the study.
All stimuli were resized to $1,280{\times}1,280$ via bilinear upsampling.
To reduce bias from luminance shifts, we normalized each stimulus by mapping its median luminance to $8$\cdms{}.

\paragraph{Apparatus and viewing conditions}
Stimuli were displayed on an ASUS ProArt Display PA32UCXR (32-inch, $3,840{\times}2,160$), a mini-LED HDR monitor supporting the Rec.~2020 color gamut.
The measured peak luminance was $1,987$\cdms{} and the minimum luminance was $<0.01$\cdms{}.
Participants viewed the display from approximately $80~\mathrm{cm}$ ($\approx$76 pixels per degree) in a darkened room.

\paragraph{Procedure} We recruited $26$ participants ($15$ males and $11$ females), aged between $20$ and $30$ years, all with normal or corrected-to-normal color vision.
We employed a pairwise comparison protocol.
On each trial, observers viewed two HDR images side by side and selected the one that appeared more natural, defined in the instructions as ``closer to a high-quality photograph of the real world.''
For text-to-HDR evaluation, prompts were not shown since our adaptation is not designed to change text-image alignment of the base T2I model. Spatial left/right placement and temporal image order were randomized, and participants took as long as needed before responding via left/right keyboard. To improve rating efficiency, we used an active sampling strategy~\cite{mikhailiuk2021active} to prioritize informative pairs; each participant completed $60$ trials for text-to-HDR and $100$ trials for RAW-to-HDR, requiring approximately $8$ and $15$ minutes, respectively.

\paragraph{Analysis}
Our pairwise comparison protocol used an active sampling strategy based on approximate message passing~\cite{mikhailiuk2021active} to reduce comparisons while maintaining accurate perceptual scaling. Pairwise outcomes were converted to JOD scores using publicly available software\footnote{\url{https://github.com/mantiuk/pwcmp}} that performs maximum likelihood estimation under the Thurstone Case V model~\cite{thurstone1994law}.
Because JOD is identifiable only up to an additive constant, we selected the origin of the JOD scale separately for each task. For RAW-to-HDR, we anchored the scale by assigning the input RAW image a JOD score of zero. As for text-to-HDR, where no natural reference condition exists, we centered the fitted scores by subtracting a constant so that the grand mean across methods is zero (\ie, a purely conventional choice of origin). With the scale fixed, we summarize each method by reporting in Figure~\ref{fig:perceptual_study} its mean JOD across stimuli, together with $95\%$ confidence intervals estimated via bootstrapping.

For text-to-HDR, \sysName\ achieves the highest perceived quality, with the \texttt{FLUX}-based variant scoring $1.81$ JOD, followed by the \texttt{SD}-based variant at $0.90$ JOD. These results indicate a consistent preference for \sysName\ over previous bracket-and-merge methods, and further suggest that backbone capacity (\texttt{FLUX.1-dev} vs. \texttt{SD-1.5}) notably impacts perceptual quality under our adaptation.
For RAW-to-HDR, \sysName\ achieves $1.87$ JOD, nearly matching the GT HDR reference. In contrast, LEDiff and RawHDR yield only statistically insignificant improvements over the RAW input, while Bracket Diffusion performs worse  with $-0.60$ JOD, which we believe is attributed to the resolution limitation under the current setting.

Per-image JOD scores are visualized in  Figure~\ref{fig:perceptual_study_supp} with $95\%$ confidence intervals.
In RAW-to-HDR, for extremely low-exposure cases (\eg, the 3rd and 4th test scenes), the HDR references contain finer textural details than our reconstructions, resulting in a relatively larger perceptual gap.
Otherwise, \sysName\ is comparable to (and sometimes perceived as better than) GT.

Overall, the perceptual results corroborate the trends in Section~\ref{sec:exp}, showing that \sysName\ consistently delivers superior visual quality for both HDR generation and reconstruction tasks.

\begin{figure*}[t]
    \centering
    \includegraphics[width=\textwidth]{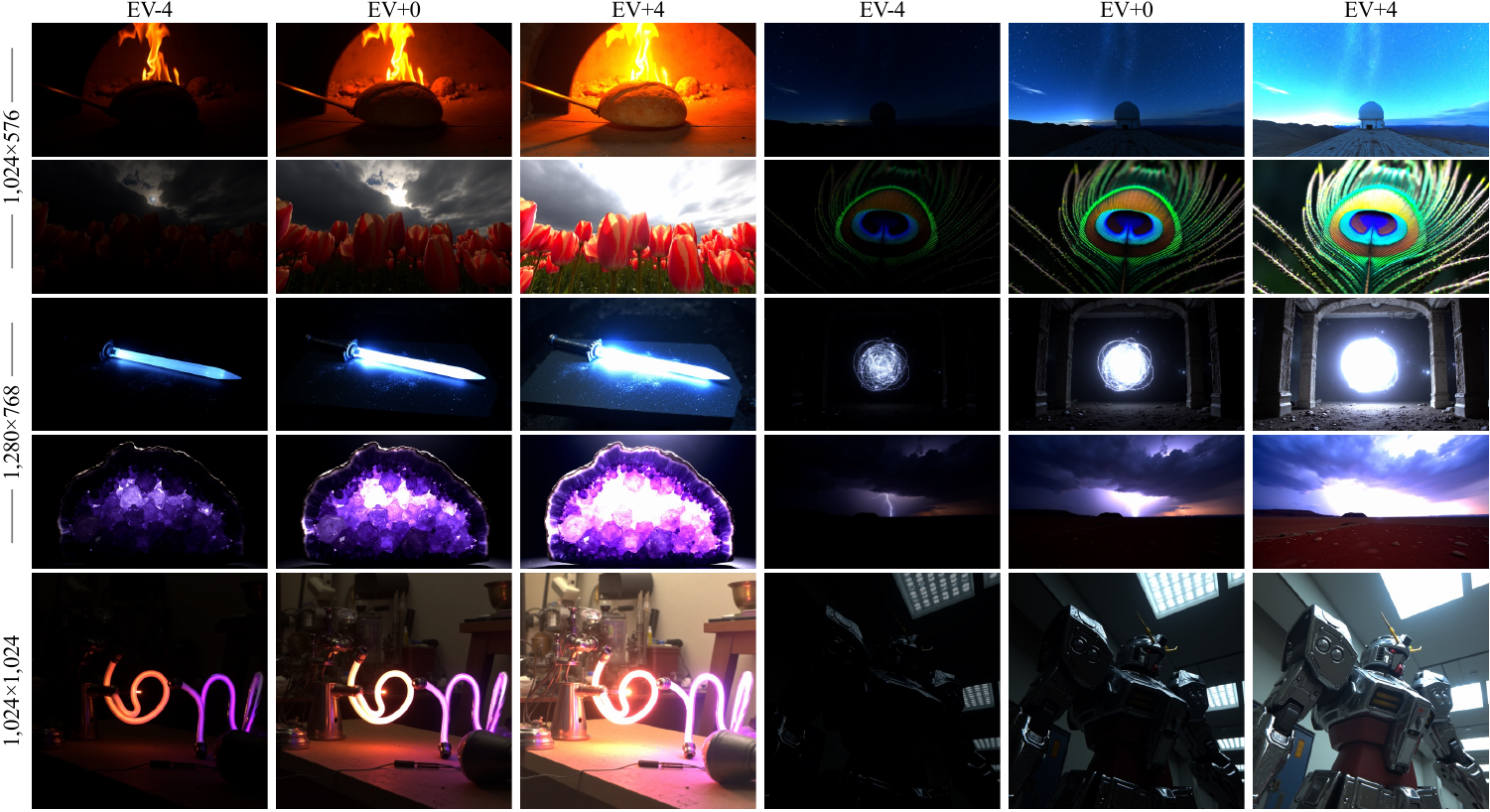}
    \caption{
    Multi-resolution HDR generation results.
    Examples are shown at three output resolutions, covering diverse lighting conditions and luminous phenomena (\eg, artificial lights, fire, sunlight, and night skies). Each HDR image is visualized at EV $-4$/$0$/$+4$ to illustrate the effective dynamic range.
    }
    \label{fig:different_resolution}
\end{figure*}

\begin{figure}[t]
    \centering
    \includegraphics[width=0.75\linewidth]{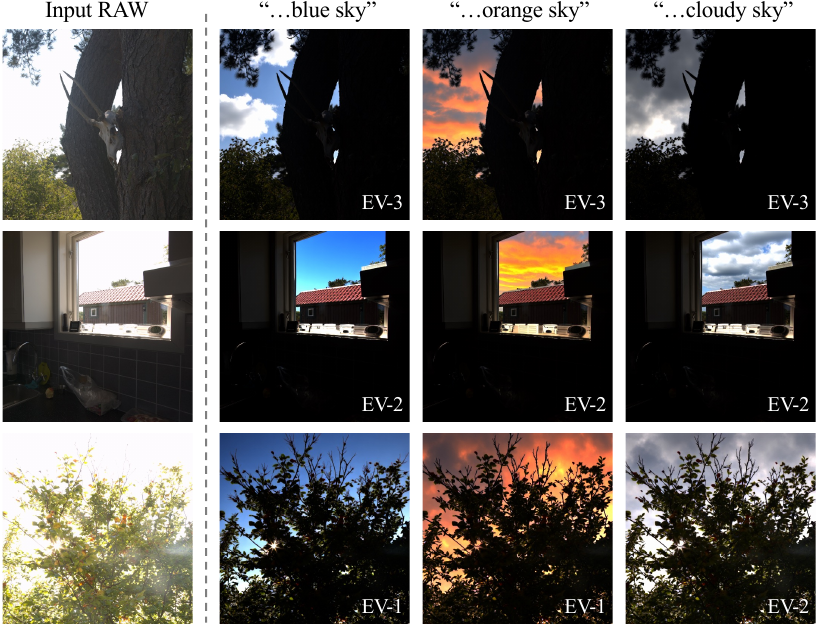}
    \caption{
    Single-RAW multi-look HDR authoring with text guidance.
    Given the same RAW input (left), our RAW-to-HDR model generates multiple HDR renderings conditioned on different prompts (\eg, ``blue sky,'' ``orange sky,'' and ``cloudy sky''), producing diverse global appearance and atmosphere while preserving scene structure.
    }
\label{fig:appl_multilook}
\end{figure}

\subsection{Higher-Resolution Results}
\label{sec_supp:different_resolution}
To ensure a fair comparison, earlier experiments fix the output resolution to  $512{\times}512$. In fact, \sysName\ is built upon modern T2I backbones and has no inherent resolution constraint. Examples in
Figure~\ref{fig:different_resolution} demonstrate text-to-HDR synthesis at $1,024{\times}576$, $1,280{\times}768$, and $1,024{\times}1,024$ across diverse lighting conditions (\eg, strong point emitters, outdoor daylight, and low-light astronomical scenes), with exposure-adjusted views showing preserved details and wide effective dynamic range. Similar generalizability extends to RAW-to-HDR reconstruction.

\subsection{Single-RAW Multi-look HDR Authoring}
\label{sec_supp:text_guided_hallucination}
Beyond ``pure'' reconstruction where text prompts are dropped, our  RAW-to-HDR model can also be used as a one-to-many HDR authoring tool: conditioning the same RAW input on different prompts produces multiple HDR renderings with distinct global appearance and atmosphere (see Figure~\ref{fig:appl_multilook}).  Conventional workflows do so by applying tone-mapping and hand-crafted color grading in display space, which offer limited control over physically meaningful highlight behavior. In stark contrast, \sysName\ performs prompt-driven tone manipulations in a physically grounded, perceptually uniform space, while preserving scene structure.

\begin{figure}[t]
    \centering
    \includegraphics[width=0.75\linewidth]{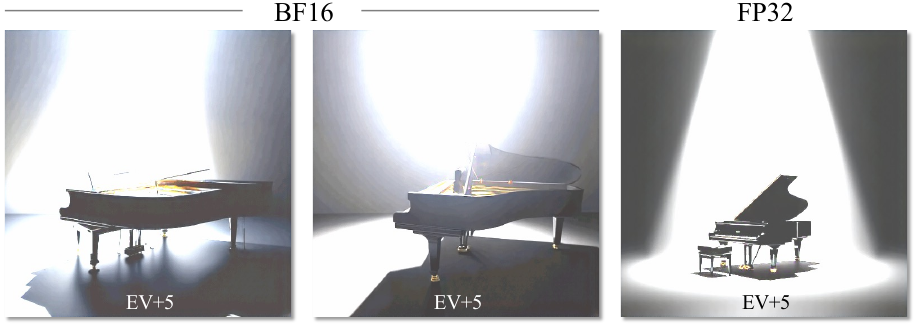}
    \caption{
    Banding artifacts induced by BF16 inference in text-to-HDR generation. When rendered at EV$+5$, quantization in smooth, low-luminance gradients produces visible banding (left/middle), whereas FP32 inference preserves subtle gradients and eliminates the artifacts (right).
    }
    \label{fig:banding_effect}
\end{figure}

\subsection{Discussion on Banding Effect}
In text-to-HDR, we occasionally observe banding artifacts in a small fraction of generated HDR images (approximately $2$ out of $100$), predominantly in dark regions with smooth luminance gradients (see Figure~\ref{fig:banding_effect}).
Under the default BF16 inference ($16$-bit representation with $8$ bits for the exponent and only $7$ bits for the mantissa), these smooth, low-luminance gradients are likely quantized into discrete bins, yielding visible banding. Using FP32 ($23$-bit mantissa) provides substantially finer resolution and effectively eliminates the artifacts. We therefore recommend FP32 inference when banding is visible, and retain BF16 as the default.

\begin{figure}[t]
    \centering
    \includegraphics[width=0.8\linewidth]{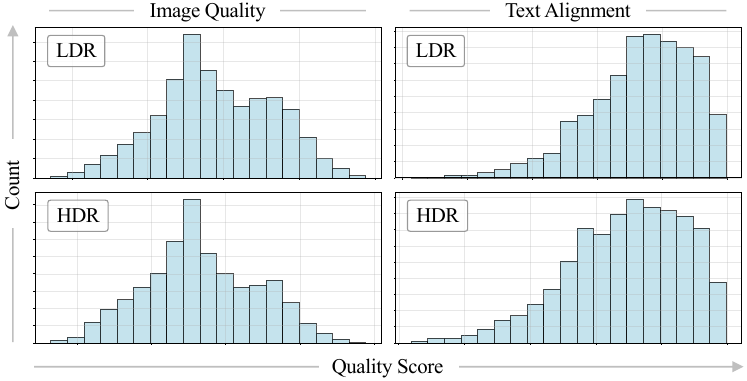}
    \caption{Q-Eval-100K score distributions for 512 aligned (LDR, HDR) pairs.
    }
\label{fig:eval_metric}
\end{figure}

\subsection{On the Use of Q-Eval-100K in Text-to-HDR}
\label{sec_supp:q_eval}
To validate the use of Q-Eval-100K~\cite{zhang2025q} for evaluating HDR image quality and text-image alignment in text-to-HDR, we conduct a sanity check on $512$ aligned (LDR, HDR) pairs.
Captions are generated by \texttt{Gemini-2.0-Flash}.
All HDR images are rescaled to $L_{\mathrm{peak}} = 4{,}000$\cdms{} and PU21-encoded before evaluation. Scores for LDR and PU21-encoded HDR images show strong cross-domain consistency: Pearson correlations are $0.933$ for image quality and $0.830$ for text-image alignment. Mean scores over the $512$ pairs are $0.611$ vs. $0.587$ (image quality) and $0.736$ vs. $0.695$ (text-image alignment) for LDR vs. PU21-encoded HDR images, respectively, suggesting reasonable transfer of Q-Eval-100K (see also the score distributions in Figure~\ref{fig:eval_metric}).

Nonetheless, we observe a consistent bias in favor of LDR images, even when matched (LDR, HDR) pairs are perceptually equivalent. Therefore, Q-Eval-100K is not reported in settings that mix LDR and HDR outputs, where it could be misleading.

\begin{figure*}[t]
    \centering
    \includegraphics[width=\textwidth]{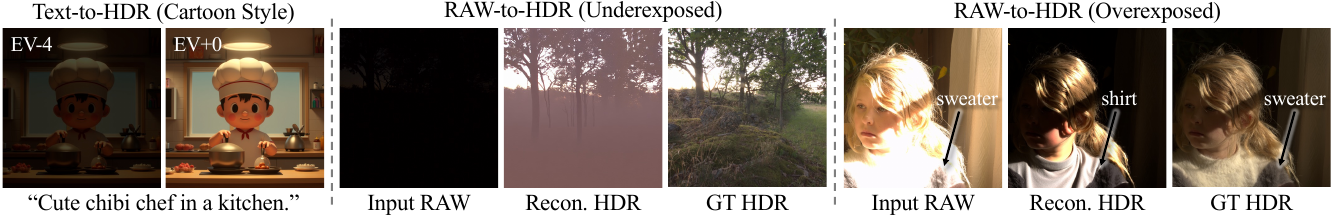}
    \caption{
    Limitations. Text-to-HDR: performance may degrade for out-of-domain styles, leading to less realistic appearance.
    RAW-to-HDR: implausible or inconsistent hallucinations may occur for large regions that are severely underexposed or overexposed.
    }
\label{fig:limitation}
\end{figure*}

\section{Limitations}
\label{sec_supp:limitations}
For text-to-HDR, our current training data are dominated by natural photographs, so the resulting model may generalize less reliably to out-of-domain styles (see the cartoons in Figure~\ref{fig:limitation}).
For RAW-to-HDR, X2HDR can still fail in extremely underexposed or overexposed regions, occasionally introducing implausible hallucinations or local detail inconsistencies (see also Figure~\ref{fig:limitation}).
In addition, the current \sysName\ is display-agnostic, \ie, it does not condition on target peak luminance or other device characteristics, and provides only limited support for controllable HDR image generation and interactive HDR editing.

\section{Future Work}
\label{sec_supp:future_work}
First, for text-to-HDR, our training data are dominated by natural scenes, which can reduce reliability to out-of-domain styles (\eg, cartoons in Figure~\ref{fig:limitation}). Style-robust finetuning and data augmentation, such as stylized HDR assets, synthetic relighting, and domain-mixing curricula, may improve generalization while preserving HDR plausibility. For RAW-to-HDR, failures in extremely underexposed or overexposed regions sometimes manifest as implausible or inconsistent local structure. This could be mitigated by incorporating perceptual objectives (\eg, DISTS), locality-aware constraints, and uncertainty- or saturation-aware conditioning that explicitly prevents implausible extrapolation.

Second, because \sysName\ adapts pretrained T2I models to HDR synthesis by LoRA finetuning, it is inherently modular and can be combined with complementary advances in the T2I ecosystem. A natural next step is to integrate \sysName\ with controllable generation techniques (\eg, ControlNet~\cite{zhang2023adding}) to unlock controllable HDR synthesis, and with emerging instruction- and context-based image editing models (\eg, \texttt{FLUX.1-Kontext}) to support HDR-aware editing.  Beyond the RAW-to-HDR setting studied here, our idea can be extended to single-image LDR-to-HDR reconstruction by conditioning on LDR inputs, and to multi-exposure inputs for HDR fusion and reconstruction.

Third, extending \sysName\ from still images to videos is an important direction. Applying the same perceptually uniform encoding and adaptation strategy to video diffusion models could spur text-to-HDR video generation and RAW-to-HDR video reconstruction, with explicit mechanisms for temporal consistency (\eg, motion-aware conditioning, recurrent features, or temporal regularizers) to prevent flickering and exposure instability. Meanwhile, conditioning generation on target display capabilities (\eg, peak luminance, local-dimming behavior, and color volume)
and incorporating power/thermal constraints during generation is of practical importance, which encourages content that ``spends'' dynamic range where it is most perceptually effective while respecting energy budgets.

Fourth, progress is currently constrained by the limited availability of large-scale, publicly accessible HDR data. Although our experiments suggest that freezing the pretrained VAE is sufficient for strong performance, future work could investigate VAE finetuning or distillation using larger HDR corpora to further improve reconstruction fidelity, particularly in extreme luminance ranges, where quantization and saturation effects are most salient.

Finally, HDR generation would benefit from more HDR-native evaluation and optimization. We currently use Q-Eval-100K~\cite{zhang2025q} as a practical proxy for no-reference HDR assessment, but such evaluators are not designed for HDR-specific perceptual factors (\eg, highlight naturalness, extreme luminance behavior, and wide-range local contrast). Developing dedicated no-reference HDR quality models is therefore a high-impact direction, unlocking scalable benchmarking, training-time filtering, and reinforcement-learning-style optimization.


\end{document}